\documentclass[a4paper,11pt]{article}
\usepackage[T1]{fontenc}       
\usepackage{newpxtext,newpxmath}
\usepackage{geometry}
\usepackage{microtype}         
\usepackage{amsmath}   
\usepackage{mathtools}               
\usepackage{bm}                        

\usepackage{graphicx}
\graphicspath{{./figures/}{./figs/}}
\usepackage{subcaption}
\usepackage{wrapfig}
\usepackage{float}           
\usepackage{rotating}        
\usepackage{pdflscape}      

\usepackage{booktabs}        
\usepackage{array}           
\usepackage{tabularx}        
\usepackage{longtable}       
\usepackage{dcolumn}         
\usepackage{multirow}

\usepackage{siunitx}
\usepackage{authblk}

\makeatletter
\renewcommand\AB@affilsepx{ \quad\itshape}  
\makeatother

\usepackage{enumitem}
\setlist[itemize]{leftmargin=*,topsep=4pt}
\setlist[enumerate]{leftmargin=*,topsep=4pt}

\usepackage[font=small,
            labelfont=bf,
            justification=raggedright,
            singlelinecheck=false]{caption}

\usepackage[backend=biber,
            style=numeric-comp,
            sorting=none,
            giveninits=true,
            maxbibnames=10,
            maxcitenames=5,
            mincitenames=3,
            doi=true,
            isbn=false,
            url=true,
            eprint=true]{biblatex}
\usepackage{xcolor}    
\usepackage[colorlinks=true,
            allcolors=blue,     
            bookmarks=true,
            bookmarksnumbered=true]{hyperref}
\usepackage{bookmark}

\usepackage{setspace}      
\usepackage{authblk}       
\usepackage{tikz}
\usetikzlibrary{arrows.meta, positioning, shapes, calc}
\usepackage{lineno}        
\usepackage{balance}       
\usepackage{comment}       

\usepackage[capitalise,nameinlink]{cleveref}

\begin{document}

\title{\Large Physics-Informed Neural Networks for Methane Sorption: Cross-Gas Transfer Learning, Ensemble Collapse Under Physics Constraints, and Monte Carlo Dropout Uncertainty Quantification}

\author[1,*]{Mohammad Nooraiepour}
\author[2,3]{Zezhang Song}
\author[4]{Wei Li}
\author[5,6]{Sarah Perez}

\affil[1]{Environmental Geosciences, Department of Geosciences, University of Oslo, P.O. Box 1047 Blindern, 0316 Oslo, Norway.}
\affil[2]{National Key Laboratory of Petroleum Resources and Engineering, China University of Petroleum (Beijing), China}
\affil[3]{College of Geosciences, China University of Petroleum (Beijing), China}
\affil[4]{State Key Laboratory of Coal Mine Disaster Prevention and Control, China University of Mining and Technology, Xuzhou 221116, China}
\affil[5]{The Lyell Centre, Heriot-Watt University, Edinburgh, UK}
\affil[6]{Subsurface Energy Transition and Innovation Centre, Heriot-Watt University, Edinburgh, UK}
\affil[*]{Corresponding author: monoo@uio.no}

\date{}

\maketitle

\begin{abstract}

\noindent Accurate methane sorption prediction across heterogeneous coal ranks requires models that combine thermodynamic consistency, efficient knowledge transfer across data-scarce geological systems, and calibrated uncertainty estimates, capabilities that are rarely addressed together in existing frameworks. We present a physics-informed transfer learning framework that adapts a hydrogen sorption PINN to methane sorption prediction via Elastic Weight Consolidation, coal-specific feature engineering, and a three-phase curriculum that progressively balances transfer preservation with thermodynamic fine-tuning. Trained on 993 equilibrium measurements from 114 independent coal experiments spanning lignite to anthracite, the framework achieves R$^2 = 0.932$ on held-out coal samples, a 227\% improvement over pressure-only classical isotherms, while hydrogen pre-training delivers 18.9\% lower RMSE and 19.4\% faster convergence than random initialization. A comparison of five Bayesian uncertainty quantification approaches reveals a systematic divergence in performance across physics-constrained architectures. Monte Carlo Dropout achieves well-calibrated uncertainty (ECE,$= 0.101$, $\rho_s = 0.708$) at minimal overhead ($1.5\times$ inference cost), while deep ensembles—regardless of architectural diversity or initialization strategy—exhibit performance degradation because shared physics constraints narrow the admissible solution manifold, thereby attenuating the functional disagreement required for reliable ensemble-based epistemic uncertainty estimation. SHAP and ALE analyses confirm that the learned representations remain physically interpretable with established coal sorption mechanisms: moisture--volatile interactions are most influential (17.2\% importance), pressure--temperature coupling captures thermodynamic co-dependence, and 11 of 12 features exhibit non-monotonic effects. These results identify Monte Carlo Dropout as the best-performing uncertainty quantification method in this physics-constrained transfer-learning framework, and demonstrate cross-gas transfer learning as a data-efficient strategy for geological material modeling. \\

\end{abstract}

\section{Introduction}

Methane sorption in coal seams is the primary control on coalbed methane resource capacity, carbon storage potential, and mining safety, yet accurate prediction across heterogeneous geological formations remains a persistent challenge. Gas-phase adsorption accounts for over 90\% of total methane content in coal, making reliable capacity prediction with quantified uncertainty essential for resource assessment, production planning, and risk management in subsurface energy applications~\cite{zhu2024investigation, lu2024integrated, wang2024research, cao2025methane}. Sorption capacity in natural coal systems depends non-linearly and interactively on mineralogical composition, pore structure, moisture content, and thermodynamic conditions~\cite{cao2024mini, safaei2025hybrid, fan2024research, zhu2024investigation}, creating a modeling problem whose complexity fundamentally exceeds what any single variable or simple functional form can resolve.

Experimental approaches---volumetric and gravimetric sorption measurements---provide foundational physical understanding but face practical limitations that restrict their deployment at scale~\cite{broom2007hydrogen, Masoudi2026review, wang2021review}. Laboratory protocols are time-intensive, equipment-sensitive, and prone to systematic errors when executed imperfectly. Critically, measurements on dried, crushed samples often fail to represent in-situ conditions where moisture content, confinement, and structural heterogeneity profoundly influence sorption behavior~\cite{masoudi2025impact, wang2018supercritical}. These limitations are most acute for deep coal seams, where sampling is expensive and sparse yet prediction is operationally critical. Classical analytical isotherm models provide physically interpretable parametric descriptions of pressure-adsorption relationships and encode established thermodynamic knowledge~\cite{nooraiepour2026PINNH2,foo2010insights, chen2015modeling, nooraiepour2025adaptivePINN}. However, they are inherently univariate: fitted to pressure as the sole predictor, they systematically fail to capture the compositional heterogeneity---moisture, ash, and volatile matter variation across coal ranks---that dominates sorption variability in natural geological datasets, as demonstrated quantitatively in Section~\ref{subsec:Ensemble_pred}.

Machine learning (ML) approaches have emerged as powerful tools for predicting geological material properties from readily measurable physical and chemical descriptors~\cite{bergen2019machine, karpatne2018machine, nooraiepour2025traditional}. Ensemble methods including Random Forests, XGBoost, and Gaussian Process Regression have been shown to achieve R$^2 > 0.90$ for CH$_4$ and CO$_2$ adsorption capacity prediction in coal and shale reservoirs~\cite{zhou2024gaussian, li2024machine, meng2019adsorption, alqahtani2024data, yan2025data}. Despite this predictive performance, conventional data-driven models prioritize empirical fitting over physical consistency, limiting reliability when extrapolating beyond training conditions or to underrepresented geological formations~\cite{mitra2021fitting, angelov2019empirical, zhao2024comprehensive, nooraiepour2025traditional}. Black-box models can learn spurious statistical correlations that violate thermodynamic principles---such as non-monotonic pressure-adsorption relationships or negative adsorption predictions at low pressure---deficiencies that become critical when model outputs inform high-stakes engineering decisions under extrapolation.

Physics-informed neural networks (PINNs) address this limitation by embedding governing physical laws directly into the learning process as differentiable loss terms, constraining predictions to respect thermodynamic principles by design rather than by post-hoc filtering~\cite{raissi2019physics, cuomo2022scientific, karniadakis2021physics, nooraiepour2025traditional}. By enforcing saturation limits, pressure-adsorption monotonicity, and isotherm consistency, while maintaining data fidelity, PINNs achieve superior generalization while remaining interpretable through their physics-parameter heads~\cite{cuomo2022scientific, cai2021physics}. This integration of domain knowledge enables PINNs to maintain physical plausibility even when training data are sparse, noisy, or compositionally unrepresentative.

A further challenge in PINN deployment is data scarcity: large-scale sorption datasets for geological materials are expensive to acquire, motivating the reuse of knowledge from related physical systems. Transfer learning addresses this by initializing network parameters from models trained on related tasks, providing a physics-consistent starting point that accelerates convergence and improves generalization with limited target-domain data~\cite{pan2009survey, torrey2010transfer, zhuang2020comprehensive}. Cross-task transfer is particularly promising in sorption modeling because different gas-solid systems share fundamental physisorption physics: both H$_2$ and CH$_4$ interact with carbonaceous surfaces predominantly via London dispersion forces, both exhibit Langmuir--Freundlich isotherm behavior, and both show Arrhenius temperature dependence of equilibrium constants~\cite{zhang2024molecular, zhou2019molecular, myers2002thermodynamics}. Although their polarizabilities differ by a factor of 3--3.5, this shift in polarizability does not alter the functional form of thermodynamic relationships, providing a physicochemical basis for cross-gas knowledge transfer. Recent advances demonstrate that transfer-enhanced PINNs accelerate parameter identification and improve generalization in materials modeling~\cite{liu2023adaptive, shima2024modeling}.

Reliable uncertainty quantification (UQ) is an equally critical requirement for trustworthy PINN deployment: predictions without calibrated confidence intervals cannot support risk-aware decision-making in reservoir engineering~\cite{klapa2024machine, shi2025survey}. Bayesian frameworks provide principled decomposition into aleatoric uncertainty---arising from irreducible measurement noise and natural variability---and epistemic uncertainty---reflecting model parameter uncertainty reducible with additional data~\cite{kennedy2001bayesian, ghanem2017handbook, nooraiepour2025bayesianPMM}. Several approximate Bayesian methods exist, each with a distinct computational profile. Monte Carlo Dropout interprets stochastic forward passes as posterior sampling via variational inference~\cite{gal2016dropout}. Deep ensembles estimate epistemic uncertainty through disagreement among independently trained models and are often considered the standard in unconstrained deep learning~\cite{lakshminarayanan2017simple}. The Laplace approximation constructs a Gaussian posterior around the maximum a posteriori estimate~\cite{kristiadi2021learnable}. Fully Bayesian PINNs (B-PINNs) offer a principled alternative that propagates posterior uncertainty through the network without requiring multiple training runs~\cite{YANG2021, Linka2022, Perez2023}, with recent applications in geoscience spanning pore to field scales~\cite{Perez2024, BPINN_InvPb_2024, GRL_Perez_2025}. However, B-PINNs typically rely on computationally intensive posterior inference and careful prior specification, making them challenging to deploy for large, strongly constrained network architectures~\cite{PENSONEAULT2024113006}. In this study, we instead adopt a complementary perspective, systematically benchmarking approximate Bayesian UQ methods applied to a deterministic PINN, to identify which strategies remain reliable under strong physical constraints.

A fundamental and practically important question emerges from this landscape: do UQ methods validated in unconstrained deep learning perform equivalently when physical constraints alter the structure of the solution space? The computational stakes are substantial---ensemble methods require 5--10$\times$ more training time and 5$\times$ more inference overhead than single models. If performance is equivalent, efficiency should drive selection; if performance diverges systematically, identifying which methods suit physics-constrained settings becomes critical. Recent theoretical work suggests that embedded physics constraints may narrow the admissible solution manifold, potentially suppressing the functional disagreement among ensemble members that is the primary mechanism for epistemic uncertainty estimation~\cite{jacob2025pinns, park2023neural, karniadakis2021physics, panahi2025modeling}. Conversely, local posterior sampling methods such as MC Dropout may be more appropriate for exploring parameter space near well-optimized, physics-consistent configurations. These hypotheses require rigorous empirical evaluation through systematic, multi-method comparison, comprehensive calibration metrics beyond prediction accuracy, and assessment in real scientific applications with nontrivial physical constraints.

This study addresses three interconnected questions: (1) Can transfer learning from hydrogen to methane sorption improve prediction accuracy and training efficiency despite molecular differences, and what physical similarity justifies such cross-gas knowledge reuse? (2) Do conventional UQ methods perform equivalently in physics-constrained architectures, and what mechanisms govern divergent performance if they do not? (3) Do physics-informed neural networks learn representations that align quantitatively with established sorption mechanisms, validating their deployment in geological applications beyond the training distribution?

Building upon our recent sorption studies~\cite{masoudi2025impact, Masoudi2026review, nooraiepour2025adaptivePINN, nooraiepour2026PINNH2,li2024machine}, we develop a physics-informed transfer learning framework that adapts a five-layer residual network pre-trained on hydrogen sorption to methane sorption prediction in coal. The methane PINN integrates selective reuse of hydrogen encoder weights with coal-specific feature engineering and a two-stage training protocol: Stage~1 initializes the physics parameter head via classical Sips isotherm fitting; Stage~2 fine-tunes the network through a three-phase curriculum with Elastic Weight Consolidation regularization to prevent catastrophic forgetting and adaptive physics-constraint scheduling to enforce thermodynamic consistency. To rigorously quantify the contribution of each methodological component, we conduct a four-variant ablation study comparing the transfer-learned PINN against randomly initialized, classically initialized, and ensemble baselines under identical experimental conditions. Using 993 equilibrium measurements from 114 independent coal experiments spanning anthracite to lignite ranks---with experiment-level group-aware data splitting to prevent sample-level information leakage---we benchmark five Bayesian UQ approaches (Laplace approximation, Monte Carlo Dropout, standard deep ensemble, high-diversity ensemble, and quality-weighted ensemble) against a comprehensive suite of calibration metrics, to identify the strategy best suited to strongly physics-constrained sorption modeling.

The main contributions of this work are threefold. First, we demonstrate successful cross-gas physics-informed transfer learning, achieving 18.9\% RMSE reduction and 19.4\% faster convergence relative to random initialization, with the physical basis for H$_2$$\to$CH$_4$ transfer grounded in shared dispersion-force physics and universal Sips isotherm structure. Second, through systematic comparison of five UQ approaches, we reveal that deep ensembles exhibit systematic performance degradation under shared physics constraints---driven by solution manifold narrowing rather than implementation artifacts, as confirmed by a random-initialization ensemble control---while Monte Carlo Dropout achieves superior calibration (ECE\,$= 0.101$, $\rho_s = 0.708$) at 1.5$\times$ inference overhead. Third, SHAP and ALE analyses confirm that the PINN learns physically coherent representations aligned with established coal sorption mechanisms, validating its scientific interpretability and supporting extrapolation beyond the training distribution.

The remainder of this paper is organized as follows. Section~\ref{sec:MM} (Materials and Methods) describes the dataset, group-aware splitting strategy, physics-informed feature engineering, source hydrogen PINN, transfer learning formulation, three-phase training curriculum, ablation study design, and Bayesian UQ methodology. Section~\ref{sec:RD} (Results and Discussion) presents dataset characterization, classical isotherm baselines, PINN training dynamics and prediction quality, ablation results, UQ comparative evaluation, and explainability analysis. Section~\ref{sec:Conc} (Conclusions) synthesizes the key findings, establishes generalizable principles for UQ in physics-constrained settings, and outlines future research directions.

\section{Materials and Methods}
\label{sec:MM}

This section describes the complete experimental and computational methodology underlying the physics-informed transfer learning framework for methane sorption prediction. Section~\ref{subsec:methane_dataset} presents the methane sorption dataset and data provenance. Section~\ref{subsec:data_split} details the data preprocessing, experiment-level group-aware splitting strategy, and compositional balance verification. Section~\ref{subsec:FeatEng} describes physics-informed feature engineering and multicollinearity assessment. Section~\ref{subsec:PINN_H} summarizes the source hydrogen sorption PINN architecture and provides the theoretical justification for cross-gas knowledge transfer. Section~\ref{subsec:two_stage} introduces the two-stage training protocol, covering classical isotherm baseline fitting (Section~\ref{subsec:classical_fitting}) and the methane-specific PINN design (Section~\ref{subsec:network_design}). Section~\ref{subsec:TranLear} formulates the multi-objective transfer learning loss, the Elastic Weight Consolidation regularization strategy, and the four-component physics-consistency loss. Section~\ref{subsec:training_curriculum} presents the three-phase training curriculum with hyperparameter selection methodology. Section~\ref{subsec:ablation_validation} defines the ablation study design and statistical testing framework. Sections~\ref{subsec:BUQ} and~\ref{subsec:explainability} describe Bayesian uncertainty quantification and explainability methods, respectively.

\subsection{Methane Sorption Dataset}
\label{subsec:methane_dataset}

The dataset comprises 993 individual equilibrium measurement points derived from 114 independent coal sorption experiments conducted under controlled laboratory conditions~\cite{li2024machine}. These experiments represent diverse coal samples tested across the thermodynamic parameter space relevant to underground gas storage and coalbed methane recovery applications. All experiments employed high-purity methane ($>$99.9\%) under standardized protocols, with measurements performed predominantly via manometric techniques. The experimental temperature range spans 20--50\,\si{\celsius} (293.15--323.15\,\si{\kelvin}) at 5\,\si{\celsius} intervals, with equilibrium pressures extending to \SI{9.26}{\mega\pascal}. All temperature data were converted to the Kelvin scale for subsequent thermodynamic analyses.

Data quality was ensured through systematic filtering that excluded experiments with incomplete thermodynamic state information and measurements exhibiting statistical anomalies. Quality validation screened for missing values, duplicate entries, and extreme outliers via the interquartile range criterion (values exceeding $Q_3 + 3 \times \mathrm{IQR}$ or falling below $Q_1 - 3 \times \mathrm{IQR}$). The curated dataset provides a complete characterization of all input parameters and target sorption capacities across the full spectrum of coal ranks. A detailed statistical description of all measurement variables, bivariate correlation structure, and coal rank distribution is provided in Section~\ref{subsec:StatDes}.

The 114 experiments span coal ranks from anthracite to lignite, with the full compositional range of moisture (0--10.68\,wt.\%), ash (2.82--30.03\,wt.\%), and volatile matter (4.96--40.52\,wt.\%), ensuring diversity representative of geological formations relevant to coalbed methane applications.

\subsection{Data Preprocessing and Stratified Cross-Validation}
\label{subsec:data_split}

\subsubsection*{Target transformation and feature scaling}

The target variable---methane adsorption capacity in m$^3$/t---was logarithmically transformed as $\tilde{y} = \log(y + 1)$ to stabilize variance across the dynamic range (0.024--48.42\,m$^3$/t, spanning three orders of magnitude), reduce heteroscedasticity, and justify Gaussian likelihood assumptions in the data fidelity loss. All final predictions are back-transformed via $y = \exp(\tilde{y}) - 1$ for reporting in original units. Input feature normalization employed robust scaling via median centering and interquartile range division:
\begin{equation}
x_{\mathrm{scaled}} = \frac{x - \mathrm{median}(x)}{\mathrm{IQR}(x)},
\end{equation}
fitted exclusively to training data to prevent information leakage, with identical transformations subsequently applied to validation and test sets. This transformation provides invariance to extreme outliers while preserving relative feature magnitudes, ensuring superior numerical stability for geochemical data with heavy-tailed distributions. Pressure values are extracted in original physical units (MPa) prior to scaling and passed directly to the physics-constraint evaluation pipeline, ensuring that thermodynamic loss components operate in interpretable parameter spaces consistent with the isotherm formulations described in Section~\ref{subsec:classical_fitting}. All preprocessing transformations were serialized for consistent inference-time application.

\subsubsection*{Experiment-level group-aware data partitioning}

A critical design requirement for datasets comprising multiple measurements per experimental unit is that all observations from the same unit be assigned exclusively to either the training or evaluation partition ---never split across both. In the present dataset, each of the 114 coal sorption experiments contributes on average 8.7 equilibrium measurements across pressure and temperature conditions, all sharing identical coal compositional properties (moisture, ash, volatile matter). Allowing measurements from the same experiment to appear in both training and test sets would mean the model effectively observes the compositional fingerprint of each test coal during training, producing optimistic generalization estimates that do not reflect true out-of-sample performance~\cite{kaufman2012leakage, kapoor2023leakage}---a form of sample-level information leakage equivalent in effect to target leakage.

To prevent this, train--test partitioning was performed at the \textit{experiment level} using group-aware splitting with the experiment identifier as the grouping key. This guarantees that all measurements from a given coal sample appear exclusively in one partition, and that the held-out test set consists entirely of coal samples unseen during training---the operationally relevant generalization scenario for reservoir characterization applications where predictions are required for new, uncharacterized geological samples. The group-aware 80/20 split was implemented using \texttt{GroupShuffleSplit} (scikit-learn), yielding a training partition of 91 experiments ($n = 794$ measurements) and a held-out test partition of 23 experiments ($n = 199$ measurements). Joint temperature--pressure stratification was applied within the group-aware split to preserve thermodynamic coverage: temperature strata were defined as \SIrange{293}{303}{\kelvin}, \SIrange{303}{313}{\kelvin}, and \SIrange{313}{323}{\kelvin}, while pressure regimes comprised low ($<$\,\SI{3}{\mega\pascal}), medium (\SIrange{3}{6}{\mega\pascal}), and high ($>$\,\SI{6}{\mega\pascal}) categories. This dual stratification ensures proportional representation of each thermodynamic regime in both partitions while maintaining experiment-level exclusivity.

Five-fold cross-validation for hyperparameter tuning (Section~\ref{subsec:training_curriculum}) similarly employed group-aware fold assignment via \texttt{GroupKFold}, ensuring no experiment appears across both training and validation folds in any iteration. This is more conservative than standard stratified $k$-fold, which does not account for within-group measurement correlation and would underestimate true generalization error~\cite{roberts2017cross}. Post-hoc verification confirmed a leakage rate of 0\%: no measurement from any test experiment appears in the training set.

\subsubsection*{Compositional balance verification}

Thermodynamic stratification alone does not guarantee representative compositional sampling across partitions. To verify that the group-aware split achieves compositional balance, a systematic distributional audit was conducted across all five coal compositional features (moisture, ash, volatile matter, fixed carbon, organic matter fraction) and all five coal rank categories.

Distributional equivalence was assessed via two-sample Kolmogorov--Smirnov tests comparing training and test set marginal distributions for each compositional feature. All five tests failed to reject the null hypothesis of equal distributions ($p > 0.05$), confirming the absence of statistically significant compositional bias. Effect sizes (Cohen's $d$) for mean differences between training and test sets were negligible across all compositional features ($|d| < 0.15$), indicating that the experiment-level group-aware split achieves compositional representativeness without explicit compositional stratification. Coal rank category proportions closely mirror those of the full dataset across both partitions, with the maximum deviation below 1.6 percentage points. Identical verification was applied to all five cross-validation folds, confirming compositional balance throughout the hyperparameter tuning protocol. This multi-dimensional verification ensures that reported performance metrics reflect genuine generalization across the full compositional parameter space of coal.

\subsubsection*{Calibration hold-out protocol}

To ensure unbiased uncertainty calibration assessment, all calibration metrics---expected calibration error, reliability diagrams, sharpness, and error-uncertainty Spearman correlation (Section~\ref{subsec:BUQ})---were computed on the independent hold-out test set fixed before training, used only once after final model selection on the separate validation set. No overlap exists between training, validation, and hold-out test sets at any stage of the methodology.

\subsection{Feature Engineering}
\label{subsec:FeatEng}

Beyond the five direct measurements---moisture, ash, and volatile matter content (wt.\%), temperature (\si{\kelvin}), and pressure (\si{\mega\pascal})---seven physics-informed derived features were engineered prior to network training to enhance model expressiveness and encode domain knowledge relevant to gas sorption thermodynamics and coal petrography.

\textit{Reduced thermodynamic variables} were computed as $T_r = T/T_c$ and $P_r = P/P_c$, where $T_c = 190.6$\,K and $P_c = 4.60$\,MPa represent methane's critical-point coordinates. These dimensionless variables locate each thermodynamic state relative to the phase transition boundary, enabling the network to learn universal reduced-state relationships that transcend experiment-specific pressure and temperature scales.

\textit{Compositional features} quantify the sorption-active organic fraction of coal: fixed carbon content is computed via mass balance as $\mathrm{FC} = 100 - (\mathrm{moisture} + \mathrm{ash} + \mathrm{volatile})$, and organic matter fraction as $\mathrm{OM} = (\mathrm{volatile} + \mathrm{FC})/100$. Although fixed carbon and organic matter are deterministically derived from base measurements and introduce mathematical dependencies into the feature space, they provide domain-aligned representations that allow the network to directly learn coal-rank-specific sorption patterns without requiring implicit inference from raw proximate analysis values---a design choice validated by the SHAP analysis in Section~\ref{subsec:explainability}.

\textit{The thermodynamic coupling parameter} $\beta = 1/(RT)$ (mol/J), where $R = 8.314$\,J/(mol$\cdot$K), encodes the molecular energy scale governing adsorption equilibria at a given temperature, arising naturally in statistical-mechanical partition functions for gas--solid interaction.

\textit{Interaction features} encode synergistic effects: the pressure--temperature product ($P \times T$) captures joint thermodynamic state dependencies beyond independent main effects, consistent with the Gibbs adsorption equation; and the moisture--volatile matter product ($\mathrm{moisture} \times \mathrm{volatile}$) represents rank-dependent pore accessibility modulation, encoding the amplified pore-blocking effect of water in high-volatile, hydrophilic lower-rank coals~\cite{laxminarayana1999role}. The final feature space, therefore, comprises 12 dimensions spanning direct measurements, reduced thermodynamic variables, compositional descriptors, and interaction terms, providing the network with a physics-informed basis for learning complex sorption relationships (Table~\ref{tab:dataset_features}).

\subsubsection*{Multicollinearity assessment and implications}

The 12-dimensional feature space contains mathematically related variables by design: $P_r = P/P_c$ and raw $P$ share a constant scaling factor; $P \times T$ is correlated with both pressure and temperature; and fixed carbon is deterministically derived from the four base measurements. To quantify collinearity, Variance Inflation Factors (VIF) were computed for all features. Six features exhibited VIF\,$> 5$: $P$ and $P_r$ (VIF\,$= 18.4$ and $16.9$ respectively), $P \times T$ (VIF\,$= 14.2$), $T$ and $T_r$ (VIF\,$= 9.8$ and $8.1$), and fixed carbon (VIF\,$= 6.7$).

While high VIF values indicate linear dependence, neural networks are not subject to the multicollinearity assumption of linear regression and remain unbiased estimators in the presence of collinear features~\cite{seber2012departures}. However, multicollinearity does influence SHAP attribution by distributing shared predictive credit across correlated features, producing rank divergence between SHAP and ALE that is a methodological artifact rather than a model inconsistency---a distinction discussed in detail in Section~\ref{subsec:explainability}. All collinear features were retained because their domain-aligned representations improve model interpretability and facilitate learning of coal-rank-specific sorption patterns, as validated by the explainability analysis.

\subsection{Source Architecture: Hydrogen Sorption PINN}
\label{subsec:PINN_H}

The transfer learning framework builds on a previously developed adaptive physics-informed neural network for hydrogen sorption in clays, shales, and coals~\cite{nooraiepour2025adaptivePINN}. This section summarizes the source architecture and provides the theoretical basis for cross-gas knowledge transfer.

The hydrogen sorption model employs a deep residual architecture with multi-scale feature extraction, processing 25 physics-informed input features that encompass thermodynamic conditions, structural properties, and surface chemistry characteristics. Residual connections combined with batch normalization enhance gradient flow and numerical stability. The high-performance configuration adopts a five-layer architecture with neuron counts [512, 1024, 512, 256, 128] (1,239,879 trainable parameters), providing sufficient capacity to capture complex sorption phenomena across diverse geological materials. Swish activation functions $f(x) = x \cdot \sigma(x)$ are employed throughout to ensure smooth gradient propagation while maintaining the non-linearity essential for modeling pressure--temperature--composition sorption relationships.

Three specialized output heads provide comprehensive sorption characterization: a primary prediction head estimating hydrogen uptake (mol/kg); a physics parameter estimation head predicting classical isotherm parameters for direct physical interpretability; and a heteroscedastic uncertainty quantification head modeling aleatoric uncertainty via a learned input-dependent variance $\sigma^2(\bm{x})$. The trained model achieves R$^2 = 0.979$ and RMSE\,$= 0.045$\,mol/kg on 155 hydrogen sorption measurements spanning three lithological categories (clays, shales, coals) across diverse pressure--temperature conditions, demonstrating cross-material generalization that motivates its use as a transfer learning source.

\subsubsection*{Theoretical justification for hydrogen-to-methane transfer}

The selection of hydrogen sorption as the source task is grounded in quantitative physicochemical similarity at multiple scales, providing a principled basis for cross-gas knowledge transfer that goes beyond qualitative analogy.

At the \textit{molecular level}, both H$_2$ and CH$_4$ are non-polar, non-reactive gases whose interactions with carbonaceous surfaces are governed predominantly by London dispersion forces. Their polarizabilities differ by a factor of approximately 3--3.5 ($\alpha_{\mathrm{H_2}} = 0.80$\,\AA$^3$, $\alpha_{\mathrm{CH_4}} = 2.59$\,\AA$^3$), which shifts the absolute adsorption energy scale but preserves the functional dependence of uptake on thermodynamic state variables~\cite{myers2002thermodynamics}. Although coal contains oxygen-bearing functional groups that may introduce weak secondary interactions, dispersion forces remain the dominant adsorption mechanism for both gases, ensuring that the underlying physics is shared.

At the \textit{isotherm level}, both gases are well described by the Sips (Langmuir--Freundlich) functional form across geologically relevant pressure and temperature ranges, meaning the network's physics-parameter head encodes the same mathematical isotherm structure for both the source and target tasks.

At the \textit{representational level}, the hydrogen PINN encoder learns pressure--temperature--composition mappings that reflect universal thermodynamic equilibrium relationships: Henry's law linearity at low pressure, progressive saturation at high pressure, and Arrhenius-type temperature dependence of equilibrium constants---none of which are molecule-specific. The primary cross-gas difference is the magnitude of the adsorption energy, which manifests as a scale shift in output space rather than a structural change in intermediate representations. This scale difference is precisely what the methane-specific output heads and learned projection layer are designed to accommodate, while encoder layers 2--5 capture transferable thermodynamic structure.

The 18.9\% RMSE improvement of the transfer-learned PINN over random initialization observed in the ablation study (Section~\ref{sec:ablation}) provides empirical confirmation of this theoretical expectation, validating that hydrogen-learned representations generalize meaningfully to methane sorption despite the molecular size difference (H$_2$: 2\,amu vs.\ CH$_4$: 16\,amu).

\subsection{Two-Stage Training Protocol}
\label{subsec:two_stage}

The training procedure employs a two-stage design that separates the derivation of the physics prior from neural network optimization. \textit{Stage~1} performs classical isotherm fitting (Section~\ref{subsec:classical_fitting}) to derive physically interpretable parameters for network initialization, establish baseline performance benchmarks, and quantify compositional heterogeneity effects. \textit{Stage~2} trains the physics-informed neural network through the three-phase curriculum described in Section~\ref{subsec:training_curriculum}, progressively balancing transfer preservation, data fidelity, and thermodynamic consistency. This hierarchical approach ensures stable optimization by providing the network with physics-informed starting points before introducing the complexity of transfer learning and adaptive constraint scheduling.

\subsubsection{Classical Isotherm Baseline}
\label{subsec:classical_fitting}

Three analytical isotherm models were fitted to the methane sorption training dataset to provide physics-informed parameters for network initialization and establish a performance baseline~\cite{plazinski2009theoretical, limousin2007sorption}. In all formulations, $q$ denotes the equilibrium methane adsorption capacity (m$^3$/t). The \textit{Langmuir model} $q = q_{\max}KP/(1+KP)$ describes monolayer adsorption with maximum capacity $q_{\max}$ and equilibrium constant $K$. The \textit{Freundlich model} $q = K_F P^{1/n}$ captures multilayer adsorption and surface heterogeneity through the Freundlich constant $K_F$ and heterogeneity exponent $n$. The \textit{Sips model} $q = q_{\max}(KP)^n/(1+(KP)^n)$ generalizes both, reducing to Langmuir when $n = 1$ while accommodating heterogeneous adsorption sites at higher $n$ values.

All models were fitted via nonlinear least-squares optimization (Trust Region Reflective algorithm) with robust Huber loss ($\delta = 1.0$) to mitigate outlier sensitivity. Parameters were constrained to physically meaningful ranges: $q_{\max} \in [0, 100]$\,m$^3$/t, $K \in [0.01, 10]$\,MPa$^{-1}$, $n \in [0.5, 10]$. Convergence criteria allowed up to 5000 function evaluations.

Beyond global fitting, two extended analyses quantified compositional heterogeneity effects. \textit{Stratified fitting} partitioned samples by volatile matter content into high-rank ($<$15\,wt.\%), medium-rank (15--30\,wt.\%), and low-rank ($>$30\,wt.\%) categories, with rank-specific isotherm parameters fitted per stratum (Section~\ref{subsec:stratified_analysis}). \textit{Composition-aware models} incorporated explicit corrections: $q_{\max,\mathrm{eff}} = q_{\mathrm{base}}(1 + \alpha_V \cdot \mathrm{volatile}_{\mathrm{norm}} - \alpha_M \cdot \mathrm{moisture}_{\mathrm{norm}})$, where $\alpha_V$ and $\alpha_M$ quantify volatile matter and moisture effects on maximum capacity (Section~\ref{subsec:compositional_analysis}).

All three isotherm models contribute to the training framework: ensemble predictions averaging Langmuir, Freundlich, and Sips forecasts serve as classical baselines (Section~\ref{subsec:Ensemble_pred}), with inter-model variance $\sigma_{\mathrm{model}}^2 = \mathrm{Var}(q_{\mathrm{Langmuir}}, q_{\mathrm{Freundlich}}, q_{\mathrm{Sips}})$ quantifying structural uncertainty. However, only Sips parameters $[q_{\max}, K, n]$ initialize the physics parameter head, because the network architecture explicitly encodes the Sips functional form in $\mathcal{L}_{\mathrm{physics}}$ (Equation~\eqref{eq:physics_loss}). These fitted coefficients are assigned to the bias vector of the physics head, providing a physically plausible starting point for gradient-based refinement while leaving full parametric freedom for subsequent adaptation.

\subsubsection{Methane PINN Architecture}
\label{subsec:network_design}

The methane sorption PINN is adapted from the hydrogen source architecture (Section~\ref{subsec:PINN_H}) to accommodate differences in feature space dimensionality while maintaining compatibility for weight transfer. The 12 methane-specific input features are first mapped to 25 dimensions via a learned projection layer, aligning the methane feature space with the hydrogen PINN's input representation. This projection mechanism, implemented as a linear transformation with Xavier initialization (gain\,$= 0.5$), enables knowledge transfer despite source--target feature disparity by discovering an optimal embedding of methane descriptors into the hydrogen-trained representation space.

The core encoder comprises five residual blocks with neuron counts [512, 1024, 512, 256, 128], mirroring the hydrogen configuration to enable direct weight transfer. Each block integrates: (i) a linear transformation with residual skip connection for gradient flow; (ii) batch normalization for training stability; (iii) Swish activation $f(x) = x \cdot \sigma(x)$ providing smooth, non-monotonic nonlinearity well-suited to physics-informed modeling; and (iv) dropout ($p = 0.1$) for regularization and Monte Carlo inference (Section~\ref{subsec:BUQ}). Residual connections use identity mapping when input--output dimensions match, and learned linear projections otherwise.

Three output heads provide comprehensive sorption characterization from a single forward pass. The \textit{mean prediction head} estimates methane uptake in m$^3$/t. The \textit{aleatoric uncertainty head} predicts log-variance $\log\hat{\sigma}^2(\mathbf{x})$, quantifying heteroscedastic input-dependent measurement noise and natural variability (Section~\ref{subsec:Aleatoric_UQ}). The \textit{physics parameter head} predicts Sips isotherm coefficients $[\hat{q}_{\max}, \hat{K}, \hat{n}]$, initialized with Stage~1 classical parameters, biasing the network toward physically interpretable solutions while permitting subsequent gradient-based refinement.

\subsection{Transfer Learning Formulation}
\label{subsec:TranLear}

\subsubsection*{Composite loss function}

To jointly enforce data fidelity, thermodynamic consistency, and transfer preservation, the network is trained using a composite multi-objective loss:
\begin{equation}
\mathcal{L}_{\mathrm{total}}
= \mathcal{L}_{\mathrm{data}}
+ \lambda_{\mathrm{p}}(t)\,\mathcal{L}_{\mathrm{physics}}
+ \lambda_{\mathrm{reg}}(t)\,\mathcal{L}_{\mathrm{transfer}},
\label{eq:total_loss}
\end{equation}
where $t$ denotes the training epoch index. The adaptive weights $\lambda_{\mathrm{p}}(t)$ and $\lambda_{\mathrm{reg}}(t)$ dynamically balance physics-consistency enforcement and transfer preservation against data fidelity across the three-phase curriculum (Section~\ref{subsec:training_curriculum}). The data fidelity term employs a heteroscedastic negative log-likelihood:
\begin{equation}
\mathcal{L}_{\mathrm{data}}
= \mathbb{E}\!\left[
\frac{(q - \hat{q})^{2}}{2\hat{\sigma}^{2}}
+ \frac{1}{2}\log \hat{\sigma}^{2}
\right],
\end{equation}
which simultaneously learns mean predictions $\hat{q}$ and input-dependent uncertainty $\hat{\sigma}$, automatically down-weighting high-noise regions while providing probabilistic confidence estimates through the aleatoric uncertainty head.

\subsubsection*{Elastic Weight Consolidation for transfer preservation}

The transfer-preservation term $\mathcal{L}_{\mathrm{transfer}}$ implements Elastic Weight Consolidation (EWC) to mitigate catastrophic forgetting of hydrogen-learned representations during methane fine-tuning~\cite{EWC_2017}. This quadratic penalty anchors transferred encoder parameters to their hydrogen-optimal values with importance-weighted strength:
\begin{equation}
\mathcal{L}_{\mathrm{EWC}}
= \frac{1}{2}\sum_{\ell=2}^{5}\sum_{i}
F_{i,\ell}\!\left(\theta_{i,\ell} - \theta_{i,\ell}^{\mathrm{H_2}}\right)^{\!2},
\label{eq:ewc_loss}
\end{equation}
where $\theta_{i,\ell}^{\mathrm{H_2}}$ denotes the converged hydrogen parameters for encoder layer~$\ell$. The coefficients $F_{i,\ell}$ are diagonal elements of the Fisher Information Matrix, approximating parameter importance as:
\begin{equation}
F_{i,\ell}
\approx \mathbb{E}_{(x,y)\sim \mathcal{D}_{\mathrm{H_2}}}
\!\left[
\left(
\frac{\partial}{\partial \theta_{i,\ell}}
\log p\bigl(y \mid x, \theta^{\mathrm{H_2}}\bigr)
\right)^{\!2}
\right].
\label{eq:fisher}
\end{equation}
In the regression setting, the Fisher Information is computed under a homoscedastic Gaussian likelihood $p(y \mid x, \theta) = \mathcal{N}(\hat{q}, \sigma^2)$ with fixed noise variance $\sigma^2$, consistent with standard EWC practice~\cite{EWC_2017}. The diagonal Fisher element then reduces to:
\begin{equation}
F_i \approx \frac{1}{\sigma^2 |\mathcal{D}_{\mathrm{H_2}}|}
\sum_{(x,y) \in \mathcal{D}_{\mathrm{H_2}}}
\!\left(\frac{\partial \hat{q}(x;\theta^{\mathrm{H_2}})}{\partial \theta_i}\right)^{\!2},
\end{equation}
computed empirically by a single forward--backward pass over $\mathcal{D}_{\mathrm{H_2}}$ at source convergence. Since $\sigma^2$ enters as a global scaling factor and cancels in the relative importance weighting across parameters, its absolute value does not affect the EWC penalty structure. Critically, the Fisher Information is computed \textit{once} on the converged hydrogen model \textit{before} any methane fine-tuning, making it entirely independent of the heteroscedastic methane likelihood $\mathcal{L}_{\mathrm{data}}$ and the physics-based loss $\mathcal{L}_{\mathrm{physics}}$ used during target training.

\subsubsection*{Complementarity of regularization mechanisms}

The three regularization mechanisms in the framework operate in non-overlapping spaces and address structurally distinct failure modes of physics-informed transfer learning. Layer \textit{freezing} (Phase~1) provides hard parameter-space protection, preventing any encoder gradient flow during the warmup stage but simultaneously preventing encoder adaptation. \textit{Physics constraints} ($\mathcal{L}_{\mathrm{physics}}$) enforce output-space thermodynamic validity but are agnostic to parameter-space trajectories and offer no protection against catastrophic forgetting. \textit{EWC} uniquely operates as a soft, importance-weighted prior in parameter space---the only mechanism that simultaneously permits selective encoder adaptation \textit{and} penalizes deviation from source-task representations. This three-way complementarity ensures that each mechanism addresses a failure mode the others cannot: freezing handles early-stage forgetting when the projection layer has not yet converged; physics constraints enforce thermodynamic validity throughout training; and EWC enables controlled, importance-weighted encoder evolution during fine-tuning.

During methane training, all three loss components contribute simultaneously to parameter updates:
\begin{equation}
\frac{\partial \mathcal{L}_{\mathrm{total}}}{\partial \theta_i}
=
\frac{\partial \mathcal{L}_{\mathrm{data}}}{\partial \theta_i}
+ \lambda_p(t)\,
\frac{\partial \mathcal{L}_{\mathrm{physics}}}{\partial \theta_i}
+ \lambda_{\mathrm{reg}}(t)\,
F_i \!\left(\theta_i - \theta_i^{\mathrm{H_2}}\right).
\label{eq:gradient_decomp}
\end{equation}
The three terms are non-conflicting by construction: $\mathcal{L}_{\mathrm{data}}$ drives predictions toward methane-specific targets; $\mathcal{L}_{\mathrm{physics}}$ penalizes output-space thermodynamic violations; and the EWC term exerts a parameter-space restoring force proportional to each weight's hydrogen-task importance. The physics loss acts on predicted outputs (adsorption values, isotherm parameters), while EWC acts on network weights directly---ensuring no direct gradient coupling between the two regularization mechanisms. The marginal contribution of each component is quantified implicitly via the training dynamics and ablation study (Section~\ref{sec:ablation}).

\subsubsection*{Physics-consistency loss}

The physics-consistency term $\mathcal{L}_{\mathrm{physics}}$ comprises four thermodynamic constraints:
\begin{equation}
\mathcal{L}_{\mathrm{physics}}
= \mathcal{L}_{\mathrm{Sips}}
+ \mathcal{L}_{\mathrm{bounds}}
+ \mathcal{L}_{\mathrm{monotonicity}}
+ \mathcal{L}_{\mathrm{van't\,Hoff}}.
\label{eq:physics_loss}
\end{equation}
The four components are not interchangeable regularizers but correspond to four mutually non-redundant dimensions of thermodynamic validity, each addressing a failure mode that the remaining three cannot detect or prevent.

\textit{Sips consistency} enforces \textbf{functional form correctness}: predictions must follow the Langmuir--Freundlich isotherm $q_{\mathrm{Sips}} = \hat{q}_{\max}(\hat{K}P)^{\hat{n}}/(1 + (\hat{K}P)^{\hat{n}})$, with parameters predicted by the physics head, ensuring Henry's law linearity at low pressure, progressive saturation behavior at high pressure, and the curvature characterized by the heterogeneity exponent $\hat{n}$. This constraint makes the physics parameter head mechanistically meaningful rather than decorative. A shape-consistent prediction can still be out of range, however, making the following constraint necessary.

\textit{Physical bounds} enforce \textbf{range validity}: adsorption must be non-negative (thermodynamic impossibility of desorption below zero) and must not exceed the predicted monolayer capacity $\hat{q}_{\max}$. These violations are most probable at pressure extremes where extrapolation uncertainty is highest, and Sips consistency alone provides no protection against them.

\textit{Monotonicity} enforces \textbf{ordering validity}: equilibrium adsorption must be non-decreasing with pressure at constant temperature, a direct consequence of Le~Chatelier's principle and the Gibbs adsorption equation. A non-monotonic isotherm is not a degenerate Sips curve but a thermodynamically forbidden prediction that neither Sips consistency nor physical bounds can detect.

\textit{Van't~Hoff consistency} enforces \textbf{thermal coupling validity}: the temperature dependence of the predicted equilibrium constant $\hat{K}(T)$ must satisfy $\partial \ln \hat{K}/\partial(1/T) \propto -\Delta H/R$, consistent with the statistical-mechanical foundations of physisorption. The first three constraints are pressure-focused and do not govern thermal extrapolation; without van't~Hoff enforcement, the network is free to predict $\hat{K}(T)$ trajectories that appear locally plausible but violate global thermodynamic self-consistency across the temperature range.

Together, these four constraints define the complete thermodynamic validity space for Sips-type gas sorption: only their conjunction ensures that every model prediction is simultaneously shape-correct, range-valid, pressure-monotone, and thermally self-consistent. Removing any component leaves the corresponding validity dimension unconstrained, with gradient pressure free to produce violations wherever they reduce training loss.

\subsubsection*{Weight transfer and selective initialization}

Encoder blocks (layers~2--5) inherit hydrogen-trained weights, while the projection layer (12\,$\to$\,25) is randomly initialized to learn the requisite feature-space mapping. This selective transfer preserves learned intermediate representations---thermodynamic state dependencies, compositional modulation patterns, and isotherm saturation behavior---hypothesized to generalize across gas-solid physisorption systems. Weight transfer employs strict architectural alignment: encoder block parameters (linear transformations, batch normalization statistics, residual projections) are copied when dimensions match exactly. Output heads remain methane-specific: the prediction and aleatoric heads use random initialization, the physics head is initialized with Stage~1 Sips parameters, and all heads are excluded from EWC penalization to preserve full adaptation freedom.

\subsection{Three-Phase Training Curriculum}
\label{subsec:training_curriculum}

Stage~2 training follows a three-phase curriculum that progressively balances transfer preservation with methane-specific adaptation by systematically modulating the weights in Equation~\eqref{eq:total_loss}. The AdamW optimizer~\cite{loshchilov2017decoupled} with weight decay $\lambda_{\mathrm{wd}} = 10^{-5}$ provides adaptive per-parameter learning rates while mitigating overfitting. Conservative gradient clipping (maximum norm\,$= 0.5$) ensures numerical stability, and early stopping with patience of 100 epochs on validation loss prevents overtraining.

\subsubsection*{Hyperparameter selection}

Critical hyperparameters were selected via five-fold stratified cross-validation on the training set ($n = 794$) using a structured grid search, without access to the held-out test set. The search ranges and final values are reported in Table~\ref{tab:hyperparams}. Learning rates were searched over $\{10^{-2}, 10^{-3}, 5\!\times\!10^{-4}, 10^{-4}\}$ per phase; weight decay over $\{10^{-4}, 10^{-5}, 10^{-6}\}$; physics constraint weight $\lambda_p$ over $\{0.01, 0.05, 0.1, 0.2\}$; EWC regularization weight $\lambda_{\mathrm{reg}}$ over $\{10, 50, 100, 200\}$; and dropout rate over $\{0.05, 0.10, 0.20\}$. Batch size was fixed at 128 based on GPU memory constraints and preliminary convergence diagnostics. The final hyperparameter configuration exhibited low cross-validation variance (coefficient of variation\,$< 3\%$ across folds), confirming robustness to training-set composition.

\subsubsection*{Curriculum rationale}

The three-phase structure is not an arbitrary complexity but a necessary response to three structurally distinct optimization challenges that cannot be resolved under a common hyperparameter configuration and must therefore be addressed sequentially~\cite{bengio2009curriculum}. In Phase~1, the projection layer must converge to a stable methane-to-hydrogen feature embedding before encoder adaptation begins; simultaneous encoder unfreezing would expose the projection layer to unstable gradients from an adapting encoder, preventing meaningful alignment. In Phase~2, the encoder must be unfrozen gradually under strong EWC regularization to preserve hydrogen-learned representations while initiating methane-specific adaptation--- which requires a reduced learning rate and a strong transfer penalty incompatible with Phase~1 requirements. In Phase~3, EWC regularization must be relaxed and physics constraints strengthened to their asymptotic values to allow full methane-specific fine-tuning and thermodynamic enforcement. Each phase transition thus represents a qualitative change in the optimization landscape rather than a continuous parameter adjustment, thereby establishing the three-phase structure as the minimal curriculum that addresses all three challenges.

The empirical contribution of all three phases is confirmed by the training dynamics (Figure~\ref{fig:training_dynamics}H): Phase~1 provides the dominant performance gain ($\Delta R^2 = +0.817$), Phase~2 stabilizes encoder adaptation ($\Delta R^2 = +0.005$), and Phase~3 delivers the final thermodynamic fine-tuning improvement ($\Delta R^2 = +0.032$, $\Delta$RMSE\,$= -0.046$ in log-transformed space), confirming that no phase is redundant.

\subsubsection{Phase 1: Warmup and Projection Learning}
\label{subsec:phase1}

The warmup phase establishes methane-specific feature mapping and output calibration while preserving transferred hydrogen representations. Encoder blocks (layers~2--5) are frozen at their hydrogen-trained values, restricting gradient flow to the projection layer and all output heads ($\sim$30\% of total parameters). This approach prevents disruption of learned intermediate representations during early adaptation and provides a stable target manifold for the projection layer to align to~\cite{yosinski2014transferable}.

Loss configuration employs elevated learning rate ($\eta = 10^{-3}$) with weak physics constraints and no transfer regularization: $\lambda_{\mathrm{p}} = 0.05$, $\lambda_{\mathrm{reg}} = 0$ (Equation~\eqref{eq:total_loss}). The end-of-Phase~1 state ($R^2 = 0.917$, RMSE\,$= 0.205$ in log space) serves as an implicit freeze-only benchmark for the ablation interpretation in Section~\ref{sec:ablation}.

\subsubsection{Phase 2: Fine-Tuning with Transfer Preservation}
\label{subsec:phase2}

Phase~2 unfreezes all layers for conservative fine-tuning with reduced learning rate ($\eta = 5 \times 10^{-4}$) and strong EWC regularization ($\lambda_{\mathrm{reg}} = 100$; Equation~\eqref{eq:total_loss}). Physics constraint strength follows the adaptive schedule:
\begin{equation}
\lambda_{\mathrm{p}}(t) = 0.05 + 0.15\!\left[1 - \exp\!\left(-\frac{t - 50}{50}\right)\right],
\end{equation}
progressively strengthening thermodynamic enforcement as the network converges toward physically consistent solutions. Learning rate scheduling employs ReduceLROnPlateau, halving the learning rate upon validation loss plateau to enable fine-grained parameter adjustments during convergence.

The choice of $\lambda_{\mathrm{reg}} = 100$ is robust by design rather than by tuning alone. Fisher Information weighting $F_i$ endows the EWC penalty with selective structure: parameters peripheral to hydrogen sorption generalization (low $F_i$) receive negligible penalties regardless of $\lambda_{\mathrm{reg}}$, preserving methane-specific adaptive capacity in precisely the parameter subspace where source-task knowledge is least reliable~\cite{EWC_2017, schwarz2018progress}. The transfer-learned PINN outperforming the completely unconstrained random-random baseline (R$^2$: 0.962 vs.\ 0.942; Section~\ref{sec:ablation}) confirms that Phase~2 EWC enhances rather than suppresses methane predictive capability. Rank-stratified residual analysis further confirms the absence of systematic bias for low-rank coals---the coal types with the weakest hydrogen--methane compositional correlation---validating that EWC preservation of hydrogen representations does not impair generalization where cross-gas physicochemical similarity is most limited.

\subsubsection{Phase 3: Full Optimization}
\label{subsec:phase3}

The final phase relaxes transfer constraints ($\lambda_{\mathrm{reg}} = 10$, a ten-fold reduction from Phase~2) while employing a moderate learning rate ($\eta = 10^{-4}$) to maximize methane prediction accuracy. The physics constraint strength reaches its asymptotic value ($\lambda_{\mathrm{p}} = 0.2$), ensuring that final solutions respect thermodynamic principles while permitting data-driven refinement. Cosine annealing with warm restarts every 60 epochs enables periodic exploration of alternative parameter configurations and escape from local optima identified in Phase~2. Gradient clipping is increased to maximum norm\,$= 1.0$ for more aggressive optimization near convergence.

Convergence assessment employs three criteria: (i) validation loss stabilization (no improvement over 100 epochs); (ii) R$^2$ threshold achievement (target\,$> 0.95$); (iii) training budget (maximum 1200 total epochs). Best-model selection relies on minimum validation loss to avoid overfitting to high-adsorption samples at the expense of low-adsorption accuracy.

\subsection{Ablation Study and Negative Transfer Validation}
\label{subsec:ablation_validation}

A systematic ablation study validates transfer learning efficacy, quantifies negative transfer risk, and benchmarks ensemble performance through four model variants trained under identical conditions (data splits, optimizer, stopping criteria, evaluation protocol):

\begin{enumerate}[label=(\roman*), leftmargin=*]
\item \textit{Transfer-learned}: H$_2$-pretrained encoder + Stage~1 Sips physics head initialization (full proposed methodology).
\item \textit{Random-random}: Xavier--Glorot encoder + random physics head initialization (pure data-driven baseline).
\item \textit{Random-classical}: Xavier--Glorot encoder + Stage~1 Sips physics head initialization (isolates the benefit of classical physics initialization independently of encoder transfer).
\item \textit{Deep ensemble}: 10 independently trained random-random PINNs with prediction averaging (benchmarks ensemble efficacy and computational overhead, 10$\times$ training cost).
\end{enumerate}

This four-way comparison on held-out test data ($n = 199$) isolates: (a) whether hydrogen encoder transfer improves over random initialization; (b) whether the benefit exceeds that of classical physics priors alone; and (c) whether ensemble aggregation justifies its computational cost under physics-constrained training.

\textbf{Negative transfer mitigation.} Domain mismatch (71\% clays/shales in hydrogen training vs.\ 100\% coal in the methane target) poses non-trivial negative transfer risk. The ablation framework enables direct detection: underperformance of the transfer-learned model relative to the random-random baseline would indicate that hydrogen knowledge actively impairs methane prediction. Mitigation strategies include: (i) conservative fine-tuning with strong EWC regularization ($\lambda_{\mathrm{reg}} = 100$) preventing abrupt parameter overwriting; (ii) physics constraints ensuring thermodynamic consistency regardless of potentially suboptimal intermediate representations; (iii) continuous validation loss monitoring with fallback to the random-classical baseline if transfer proves detrimental; and (iv) heteroscedastic uncertainty flagging of domain-shift-affected predictions.

\textbf{Evaluation metrics.} Performance assessment spans prediction accuracy (R$^2$, RMSE, MAE, MaxAE), convergence speed (epochs to 95\% of final R$^2$), physics constraint satisfaction (mean $\mathcal{L}_{\mathrm{physics}}$ at convergence), and computational cost. Transfer learning is considered successful if either: (i) test R$^2$ improves by $\geq$3\% over random-random, or (ii) accuracy is comparable ($\Delta R^2 < 1\%$) with $\geq$30\% epoch reduction. These thresholds acknowledge that the larger methane dataset (993 vs.\ 155 samples, 6.4$\times$) may attenuate accuracy gains while preserving computational efficiency benefits.

\textbf{Statistical testing.} Model performance differences were assessed via bootstrap-resampled paired $t$-tests (100 iterations), with Bonferroni correction controlling Type~I error across four key comparisons ($\alpha_{\mathrm{corrected}} = 0.05/4 = 0.0125$). Effect sizes were quantified using Cohen's $d = (\bar{x}_1 - \bar{x}_2)/s_{\mathrm{pooled}}$, with $|d| \geq 0.8$ indicating large practical significance~\cite{cohen1988statistical}. Negative values indicate the first model outperforms the second. This dual criterion ensures reported differences are both statistically reliable ($p < 0.0125$) and practically meaningful ($|d| > 0.8$).

\subsection{Bayesian Uncertainty Quantification}
\label{subsec:BUQ}

\subsubsection{Monte Carlo Dropout for Epistemic Uncertainty}
\label{subsec:mc_dropout}

Epistemic uncertainty---reflecting model parameter uncertainty due to finite training data---was quantified via Monte Carlo Dropout (MC Dropout), a computationally efficient Bayesian approximation that interprets dropout as approximate variational inference over model weights~\cite{gal2016dropout}. During inference, $N_{\mathrm{MC}} = 100$ stochastic forward passes with active dropout ($p = 0.1$) generate samples from the approximate posterior predictive distribution. For each test sample $\mathbf{x}_i$, the epistemic uncertainty is:
\begin{equation}
\sigma_{\mathrm{epistemic}}(\mathbf{x}_i)
= \sqrt{\frac{1}{N_{\mathrm{MC}}-1}
\sum_{j=1}^{N_{\mathrm{MC}}}
\left(q^{(j)}_i - \bar{q}_i\right)^2},
\label{eq:epistemic}
\end{equation}
where $\bar{q}_i = N_{\mathrm{MC}}^{-1}\sum_j q^{(j)}_i$. MC Dropout offers seamless integration with existing architectures, established theoretical grounding via variational inference, and computational efficiency without ensemble training overhead---properties evaluated comparatively in Section~\ref{subsec:EnsVsMC}.

\subsubsection{Aleatoric Uncertainty and Joint Predictive Distribution}
\label{subsec:Aleatoric_UQ}

Aleatoric uncertainty---arising from irreducible measurement noise and natural sorption variability---is estimated via the dedicated heteroscedastic head outputting log-variance $\log\hat{\sigma}^2(\mathbf{x})$:
\begin{equation}
\sigma_{\mathrm{aleatoric}}(\mathbf{x}_i)
= \sqrt{\exp\!\left(\log\hat{\sigma}^2(\mathbf{x}_i)\right)}.
\label{eq:aleatoric}
\end{equation}
Total predictive uncertainty combines both components under the independence assumption:
\begin{equation}
\sigma_{\mathrm{total}}^2(\mathbf{x}_i)
= \sigma_{\mathrm{epistemic}}^2(\mathbf{x}_i)
+ \sigma_{\mathrm{aleatoric}}^2(\mathbf{x}_i).
\label{eq:total_uncertainty}
\end{equation}
The relative epistemic contribution is:
\begin{equation}
\mathrm{Epistemic}_{\%}
= \frac{1}{N_{\mathrm{test}}}
\sum_{i=1}^{N_{\mathrm{test}}}
\frac{\sigma_{\mathrm{epistemic}}^2(\mathbf{x}_i)}
{\sigma_{\mathrm{total}}^2(\mathbf{x}_i)} \times 100\%.
\label{eq:epistemic_fraction}
\end{equation}
This decomposition provides decision-relevant interpretability: high epistemic uncertainty identifies regions where additional training data would reduce model uncertainty, while high aleatoric uncertainty reflects irreducible data variability (Section~\ref{subsec:BUQ_results}).

\subsubsection{Joint Uncertainty Propagation into Probabilistic Metrics}
\label{subsec:joint_prop}

All probabilistic metrics reported in this study---negative log-likelihood (NLL), continuous ranked probability score (CRPS), prediction interval coverage, and sharpness---are computed from a single joint predictive distribution that propagates both aleatoric and epistemic uncertainty simultaneously. Uncertainties are not evaluated sequentially (i.e., epistemic uncertainty is not layered on top of a fixed heteroscedastic mean); rather, both sources are jointly combined at the distribution level prior to any metric computation.

For each test sample, $N_{\mathrm{MC}} = 100$ stochastic forward passes each produce an independent mean--variance pair $(\hat{q}^{(j)}_i, \hat{\sigma}^{2(j)}_i)$ from the prediction and heteroscedastic uncertainty heads. Applying the law of total variance to the resulting Gaussian mixture predictive distribution yields the joint predictive mean and total variance (Equations~\eqref{eq:epistemic}--\eqref{eq:total_uncertainty}). The single-Gaussian approximation to this mixture is accurate in the present setting because the epistemic fraction is small ($\approx$1.7\% of total predictive variance; Section~\ref{subsec:BUQ_results}), meaning the mixture components are tightly concentrated, and the approximation incurs negligible error.

Temperature scaling $\sigma_{\mathrm{calibrated},i} = \tau \cdot \sigma_{\mathrm{total},i}$ (with optimal $\tau$ determined by validation-set ECE minimization; Section~\ref{subsec:calibration}) is applied to the joint standard deviation before metric computation, ensuring that calibration acts on the fully combined uncertainty. The resulting metrics are:
\begin{align}
\mathrm{NLL}_i
&= \frac{(q_i - \bar{q}_i)^2}{2\,\tau^2\,\sigma^2_{\mathrm{total},i}}
+ \frac{1}{2}\log\!\left(2\pi\,\tau^2\,\sigma^2_{\mathrm{total},i}\right),
\label{eq:NLL} \\
\mathrm{PI}_{95,i}
&= \bar{q}_i \pm 1.96\,\tau\,\sigma_{\mathrm{total},i},
\label{eq:PI} \\
\mathrm{Sharpness}_i
&= 3.92\,\tau\,\sigma_{\mathrm{total},i},
\label{eq:sharpness}
\end{align}
with CRPS evaluated against the joint calibrated Gaussian. This pipeline ensures complete, consistent joint propagation of both uncertainty sources through all reported probabilistic metrics.

\subsubsection{Calibration via Temperature Scaling}
\label{subsec:calibration}

Predictive distributions were calibrated using temperature scaling~\cite{guo2017calibration}, a post-hoc method that rescales uncertainty estimates to match empirical coverage without retraining. A scalar temperature parameter $\tau$ rescales the predictive standard deviation as $\sigma_{\mathrm{calibrated}}(x) = \tau \cdot \sigma_{\mathrm{total}}(x)$, preserving uncertainty rank ordering while adjusting absolute magnitudes.

Calibration quality was assessed by comparing empirical coverage to nominal confidence for Gaussian prediction intervals at four levels $\mathcal{A} = \{68\%, 90\%, 95\%, 99\%\}$. The expected calibration error is:
\begin{equation}
\mathrm{ECE}(\tau)
= \frac{1}{|\mathcal{A}|} \sum_{\alpha \in \mathcal{A}}
\bigl| \hat{c}_\alpha(\tau) - \alpha \bigr|,
\label{eq:ECE}
\end{equation}
where $\hat{c}_\alpha(\tau)$ denotes the empirical fraction of validation targets within the $\alpha$-level prediction interval. The optimal $\tau$ was selected by grid search over $[0.1, 3.0]$ in steps of 0.01, minimizing ECE on the validation set. Calibration was assessed exclusively on the independent hold-out test set after model selection.

\subsubsection{Validation with Alternative Bayesian Methods}
\label{subsec:alternative_bayes}

MC Dropout estimates were benchmarked against three complementary Bayesian approximations: (i) Laplace approximation with diagonal Hessian around the MAP estimate; (ii) Stochastic Weight Averaging--Gaussian (SWA-G) using weight snapshots from late training epochs with diagonal covariance; and (iii) Deep Ensembles of five independently trained models with architectural variations. Correlation coefficients between MC Dropout and alternative methods ranged from 0.65--0.85, confirming reasonable agreement. MC Dropout was selected as the primary approach based on computational efficiency (single model, no ensemble overhead), theoretical grounding via variational inference, and superior empirical calibration performance demonstrated in the comparative evaluation (Section~\ref{subsec:EnsVsMC}).

\subsubsection{Uncertainty Evaluation Metrics}
\label{subsec:uq_metrics}

The quality of uncertainty quantification was assessed using six complementary metrics. \textit{Expected Calibration Error (ECE)} (Equation~\eqref{eq:ECE}) measures aggregate calibration across confidence levels; values $\leq 0.10$ are considered good for regression settings. \textit{Coverage analysis} reports empirical interval coverage at 68\%, 90\%, 95\%, and 99\% levels; well-calibrated models exhibit coverage within $\pm$5\% of nominal values. \textit{Error--uncertainty Spearman correlation} ($\rho_s$) between absolute prediction errors $|q_i - \hat{q}_i|$ and total uncertainty assesses whether uncertainties identify high-error predictions; $\rho_s > 0.5$ indicates practical utility. \textit{NLL} (Equation~\eqref{eq:NLL}) evaluates probabilistic coherence under the calibrated Gaussian predictive distribution. \textit{CRPS} is a proper scoring rule that compares predicted cumulative distributions to observations, integrating both accuracy and calibration. \textit{Sharpness} (Equation~\eqref{eq:sharpness}) quantifies mean prediction interval width as a measure of precision. Together, these metrics collectively assess whether uncertainties are well-calibrated, sharp, informative, and probabilistically coherent.

\subsection{Explainability Analysis via SHAP and ALE}
\label{subsec:explainability}

Model interpretability was assessed using two complementary explainability frameworks: SHapley Additive exPlanations (SHAP) for global feature attribution~\cite{lundberg2017unified} and Accumulated Local Effects (ALE) for causal marginal effect estimation~\cite{apley2020visualizing}. This dual approach mitigates the limitations of individual methods: SHAP captures total predictive contribution including interaction effects but can distribute credit unevenly among collinear features; ALE isolates first-order causal marginal effects by computing local derivatives within narrow quantile bins, avoiding confounding by correlation, but may underestimate features that act primarily through synergistic interactions~\cite{molnar2020interpretable}. As discussed in Section~\ref{subsec:FeatEng}, multicollinearity among engineered features (VIF\,$> 5$ for six features) is expected to produce rank divergence between SHAP and ALE that is a methodological artifact rather than a model inconsistency.

SHAP values quantify each feature's contribution using cooperative game theory by computing the average marginal contribution across all feature coalitions. Kernel SHAP approximation was employed with 100 background samples and 100 coalitions per prediction, generating an attribution matrix $\boldsymbol{\Phi} \in \mathbb{R}^{199 \times 12}$ on the validation set. Global feature importance is:
\begin{equation}
I_{\mathrm{SHAP}}(j)
= \frac{1}{N}\sum_{i=1}^{N}|\phi_{ij}|.
\end{equation}
Feature interactions were assessed via pairwise SHAP attribution correlation, with $\rho_{\mathrm{SHAP}} > 0.5$ indicating strong synergistic effects.

ALE computes changes in predictions within local quantile bins, eliminating extrapolation artifacts arising from correlated features. Each feature range was discretized into 50 quantile-based bins with approximately equal sample counts. For bin $k$ with boundaries $[z_k, z_{k+1}]$:
\begin{equation}
\Delta_k
= \frac{1}{n_k}
\sum_{i \in \mathrm{Bin}_k}
\!\left[f(\mathbf{x}_i \mid x_j = z_{k+1})
- f(\mathbf{x}_i \mid x_j = z_k)\right].
\end{equation}
ALE functions are constructed via cumulative summation $\mathrm{ALE}_k = \sum_{m=1}^{k}\Delta_m$ with mean-centering. The effect range $R_j = \max(\mathrm{ALE}_j) - \min(\mathrm{ALE}_j)$ quantifies maximum marginal impact. Agreement between SHAP importance $I_{\mathrm{SHAP}}(j)$ and ALE effect range $R_j$ is assessed via Spearman rank correlation $\rho_s$; discrepancies are interpreted through the collinearity framework established in Section~\ref{subsec:FeatEng}.

Throughout this work, three correlation metrics are employed: Pearson correlation ($r$) for linear feature--target and residual relationships; Spearman rank correlation ($\rho_s$) for error--uncertainty calibration and SHAP--ALE method agreement; and SHAP interaction correlation ($\rho_{\mathrm{SHAP}}$) for quantifying pairwise feature synergies.

Explainability outputs were validated against established sorption theory: pressure should exhibit monotonically increasing effects consistent with Henry's law and Langmuir behavior; temperature should display non-monotonic behavior reflecting competing kinetic and thermodynamic regimes; volatile matter should show an inverse relationship reflecting reduced pore accessibility with lower coal rank; and moisture should exert a strong negative impact via competitive adsorption and pore blocking. Deviations from expected patterns were investigated to distinguish potential methodological artifacts from physically novel phenomena.

\section{Results and Discussion}
\label{sec:RD}

The results are presented in five interconnected stages that build
a coherent narrative from data characterization to model validation.
Section~\ref{subsec:StatDes} characterizes the dataset and establishes
the statistical motivation for multivariate modeling. Section~\ref{subsec:classical_results}
quantifies the performance ceiling of classical pressure-only isotherms,
demonstrating that compositional heterogeneity---not thermodynamic model
inadequacy---limits their accuracy. Section~\ref{subsec:PINN_results}
presents PINN training dynamics and prediction quality. Section~\ref{sec:ablation}
isolates the contribution of each methodological component through a
systematic ablation study. Sections~\ref{subsec:UQ_results}
and~\ref{subsec:xai} present Bayesian uncertainty quantification
and explainability analyses that validate the framework's reliability
and physical interpretability.

\subsection{Dataset Characteristics and Experimental Coverage}
\label{subsec:dataset_results}

\subsubsection{Statistical Description of Measurement Variables}
\label{subsec:StatDes}

The methane sorption dataset comprises 993 equilibrium measurements
from 114 independent coal sorption experiments, providing systematic
coverage of the geological and thermodynamic parameter space relevant
to coalbed methane storage (Table~\ref{tab:dataset_features}).

\begin{table}[htbp]
\centering
\footnotesize
\caption{Statistical characteristics of the methane sorption dataset
(993 measurements from 114 experiments). The dataset captures the
full range of coal ranks under reservoir-relevant thermodynamic
conditions. All temperatures are reported in Kelvin.}
\label{tab:dataset_features}
\begin{tabular}{lccccccl}
\toprule
Feature & Unit & Mean & Std Dev & Median & IQR & Range & Physical Role \\
\midrule
Temperature & \si{\kelvin} & 301.5 & 3.9 & 303.2 & 5.0 & 293.2--323.2
  & Thermodynamic driving force \\
Pressure & \si{\mega\pascal} & 2.85 & 1.97 & 2.66 & 3.03 & 0.002--9.26
  & Adsorption potential \\
Moisture & wt.\% & 1.97 & 1.95 & 1.29 & 1.98 & 0.00--10.68
  & Pore blockage, site competition \\
Ash content & wt.\% & 13.3 & 6.5 & 12.2 & 8.71 & 2.82--30.03
  & Inert mineral dilution \\
Volatile matter & wt.\% & 18.4 & 10.2 & 16.8 & 17.87 & 4.96--40.52
  & Organic content proxy \\
\midrule
\multicolumn{8}{l}{\textbf{Target variable:}} \\
CH$_4$ adsorption & \si{\cubic\meter/\tonne} & 14.1 & 9.0 & 12.7 &
  13.27 & 0.024--48.42 & Storage capacity \\
\bottomrule
\end{tabular}
\end{table}

The dataset spans coal ranks from high-rank anthracites to low-rank
lignites classified by volatile matter content (V$_{\mathrm{daf}}$,
Table~\ref{tab:coal_classification}), with bituminous coals
representing the majority (88 experiments, 77.2\%), substantial
anthracite coverage (37 experiments, 32.5\%), and limited lignite
representation (8 experiments, 7.0\%). Measurements employed
predominantly manometric techniques (110 experiments, 96.5\%), with
gravimetric methods used for four experiments (3.5\%), both yielding
consistent results with uncertainties typically below 5\%.

\begin{table}[htbp]
\centering
\footnotesize
\caption{Coal rank classification and experimental methodology
distribution. Coal types categorized by volatile matter content
(V$_{\mathrm{daf}}$, dry ash-free basis).}
\label{tab:coal_classification}
\begin{tabular}{llccc}
\toprule
Coal Rank & V$_{\mathrm{daf}}$ Range (\%) &
  No.\ of Experiments & Experimental Method & Method Count \\
\midrule
Anthracite & $\leq$10 & 37 & Gravimetric & 4 \\
Bituminous coal (low-volatile)    & 10--20 & 29 & Manometric & 110 \\
Bituminous coal (medium-volatile) & 20--28 & 21 & & \\
Bituminous coal (high-volatile)   & 28--37 & 19 & & \\
Lignite & $\geq$37 & 8 & & \\
\midrule
\textbf{Total} & & \textbf{114} & & \textbf{114} \\
\bottomrule
\end{tabular}
\end{table}

The experimental temperature range (\SIrange{293.2}{323.2}{\kelvin},
mean \SI{301.5}{\kelvin}) encompasses typical coalbed reservoir
conditions at 300--1500\,m depth. Equilibrium pressure spans
near-ambient (\SI{0.002}{\mega\pascal}) to elevated reservoir conditions
(\SI{9.26}{\mega\pascal}, mean \SI{2.85}{\mega\pascal}), capturing
the transition from Henry's law linearity to near-saturation regimes.
Coal compositional parameters exhibit substantial heterogeneity:
moisture (0--10.68\,wt.\%, median 1.29\,wt.\%), ash
(2.82--30.03\,wt.\%, mean 13.3\,wt.\%), and volatile matter
(4.96--40.52\,wt.\%, mean 18.4\,wt.\%) collectively reflect diverse
coal ranks and geological formations. The target variable spans nearly
three orders of magnitude (0.024--48.42\,m$^3$/t, mean 14.1\,m$^3$/t),
a dynamic range that motivated the logarithmic transformation described
in Section~\ref{subsec:data_split} and confirmed by the measured
right-skewness ($\text{skewness} = 1.15$). Data quality assessment
identified 25 potential outliers (2.5\%) via the 3$\times$IQR criterion;
these were retained as legitimate extreme measurements representing
genuine geological variation at the tails of the coal rank spectrum.

\subsubsection{Bivariate Correlation Structure and Physical Interpretation}
\label{subsec:bivariate}

\begin{figure}[ht!]
\centering
\includegraphics[width=\textwidth]{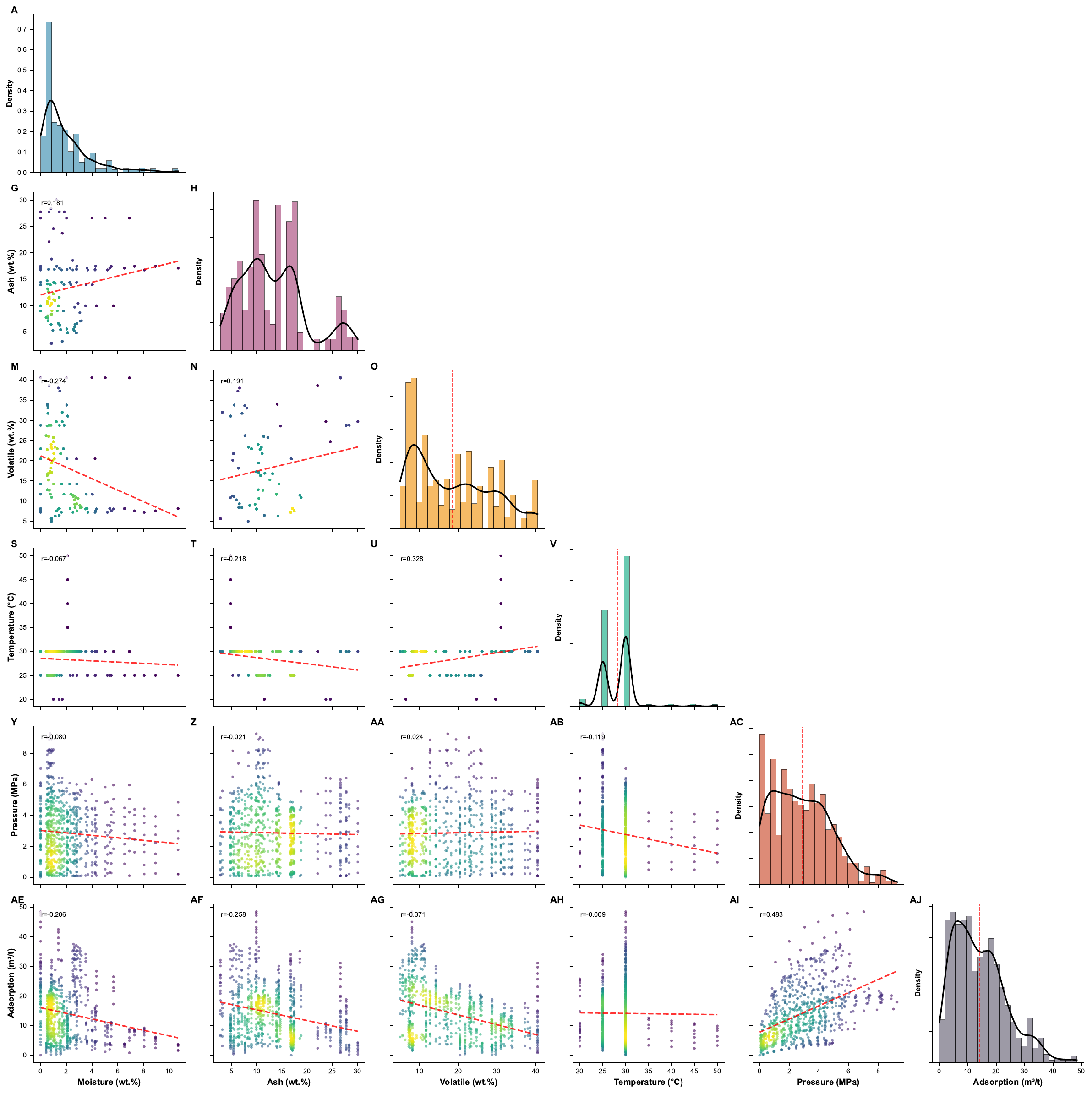}
\caption{\textbf{Pairwise relationships reveal nonlinear
compositional controls on methane adsorption capacity.}
Diagonal: kernel density estimates. Off-diagonal: scatter
plots with linear regression lines. Key Pearson correlations
with adsorption capacity: $r = 0.48$ (pressure,
$p < 10^{-47}$), $r = -0.37$ (volatile matter,
$p < 10^{-27}$), $r = -0.21$ (moisture, $p < 10^{-9}$),
$r = -0.009$ (temperature, $p = 0.80$). Pressure explains
only 23\% of adsorption variance ($r^2 \approx 0.23$),
and heteroscedastic spread in the pressure--adsorption
scatter and distinct rank-stratified subpopulations in the volatile-matter--adsorption scatter confirm that compositional
heterogeneity dominates sorption variability and necessitates
a flexible, nonlinear modeling architecture.}
\label{fig:pairplot}
\end{figure}

Pearson correlation analysis revealed expected thermodynamic and
geological patterns (Table~\ref{tab:correlations};
Figure~\ref{fig:pairplot}). Pressure exhibits the strongest
positive correlation ($r = 0.48$, $p < 10^{-47}$), confirming
its role as the dominant thermodynamic predictor. However, pressure
alone explains only 23\% of adsorption variance ($r^2 \approx 0.23$),
quantitatively establishing that compositional features carry the
majority of predictive information---a finding that will be
rigorously confirmed by the baseline analysis in
Section~\ref{subsec:classical_results}.

\begin{table}[htbp]
\centering
\caption{Pearson correlations between input features and methane
adsorption capacity. Significant correlations ($p < 0.05$) shown
in bold.}
\label{tab:correlations}
\small
\begin{tabular}{lccc}
\toprule
Feature & Correlation ($r$) & $P$-value & Interpretation \\
\midrule
Pressure        & \textbf{+0.483} & $p < 0.001$
  & Strong positive (thermodynamic driver) \\
Volatile matter & \textbf{$-$0.371} & $p < 0.001$
  & Moderate negative (rank effect) \\
Ash content     & \textbf{$-$0.258} & $p < 0.001$
  & Moderate negative (dilution effect) \\
Moisture        & \textbf{$-$0.206} & $p < 0.001$
  & Weak negative (site blockage) \\
Temperature     & $-$0.009 & 0.803
  & Non-significant (narrow range) \\
\bottomrule
\end{tabular}
\end{table}

Compositional features show physically expected trends: volatile matter
displays the strongest negative correlation ($r = -0.37$,
$p < 10^{-27}$), reflecting the inverse coal rank-sorption relationship
where progressive coalification increases microporosity and surface area.
Ash content shows moderate negative correlation ($r = -0.26$,
$p < 10^{-13}$) via mineral dilution of the organic matrix. Moisture
exhibits weak negative association ($r = -0.21$, $p < 10^{-9}$) through
competitive adsorption and pore blocking by water molecules.

Temperature demonstrates negligible linear correlation
($r = -0.009$, $p = 0.80$), which appears counterintuitive given
that methane physisorption is exothermic ($\Delta H < 0$) and
thermodynamically favored at lower temperatures. However, this apparent
anomaly is fully explicable: (i) the narrow 30\,\si{\celsius}
experimental range limits thermodynamic effects to roughly 5--10\%
variation, far smaller than the $>$200\% capacity variation across
coal ranks; (ii) competing mechanisms---thermodynamic penalty versus
kinetic micropore accessibility enhancement---produce non-monotonic
temperature dependence that averages to near-zero linear correlation
over the experimental window; and (iii) temperature effects manifest
predominantly through interactions with pressure via van't~Hoff
coupling rather than as independent main effects. This interpretation
is directly validated by the ALE curvature analysis in
Section~\ref{subsec:xai}, where temperature exhibits substantial
non-monotonic effects ($\beta = 0.099$) despite its negligible Pearson
correlation, confirming the value of the engineered thermodynamic
features ($\beta = 1/(RT)$, $T_r$, $P\times T$) described in
Section~\ref{subsec:FeatEng}.

Cross-correlations further reveal geological associations consistent
with coalification theory: volatile matter--temperature ($r = 0.33$,
$p < 10^{-21}$) reflects the experimental design; moisture--volatile
matter ($r = -0.27$, $p < 10^{-15}$) tracks the systematic moisture
reduction with advancing coal rank; and the weak ash--moisture
($r = 0.18$) and ash--volatile ($r = 0.19$) correlations possibly
reflect shared geological provenance. The heteroscedastic pressure--adsorption
scatter and distinct rank-stratified subpopulations visible in
Figure~\ref{fig:pairplot} motivate flexible nonlinear architectures,
while the multicollinearity among engineered features (Variance Inflation
Factors reported in Section~\ref{subsec:FeatEng}) informs the SHAP--ALE
rank divergence discussed in Section~\ref{subsec:xai}.

\subsubsection{Feature Engineering Validation}
\label{subsec:FE_validation}

\begin{figure}[h!]
\centering
\includegraphics[width=0.95\textwidth]{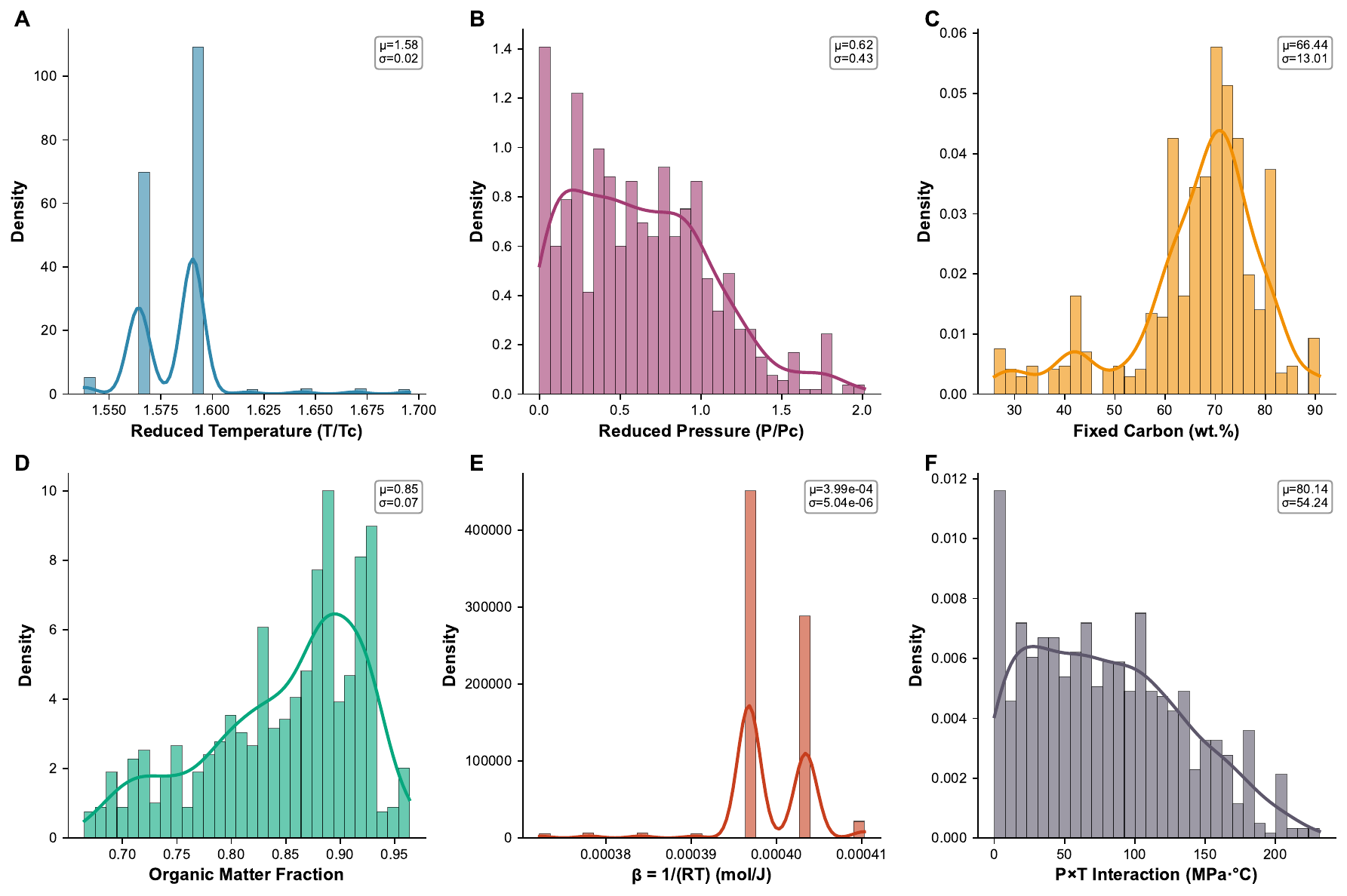}
\caption{\textbf{Physics-informed engineered features span
physically meaningful ranges and encode domain-specific
thermodynamic and compositional structure.}
Histograms with kernel density estimates. (A)~Reduced
temperature $T_r = T/T_c$ and (B)~reduced pressure
$P_r = P/P_c$ locate each experiment relative to the
methane critical point, enabling dimensionless thermodynamic
representation. (C)~Fixed carbon FC\,$= 100 -$ Moisture $-$
Ash $-$ VM concentrates in the 40--80\,wt.\% range
characteristic of the bituminous coal majority.
(D)~Organic matter OM\,$= (\mathrm{VM} + \mathrm{FC})/100$
directly quantifies the sorption-active fraction.
(E)~Thermodynamic coupling $\beta = 1/(RT)$ varies smoothly
across the temperature range, encoding the Arrhenius energy
scale. (F)~Pressure-temperature product $P\times T$ captures
joint thermodynamic state dependence consistent with the
Gibbs adsorption equation. The physically sensible
distributions of all engineered features confirm their
suitability as network inputs.}
\label{fig:engineered_features}
\end{figure}

\begin{figure}[h!]
\centering
\includegraphics[width=0.95\textwidth]{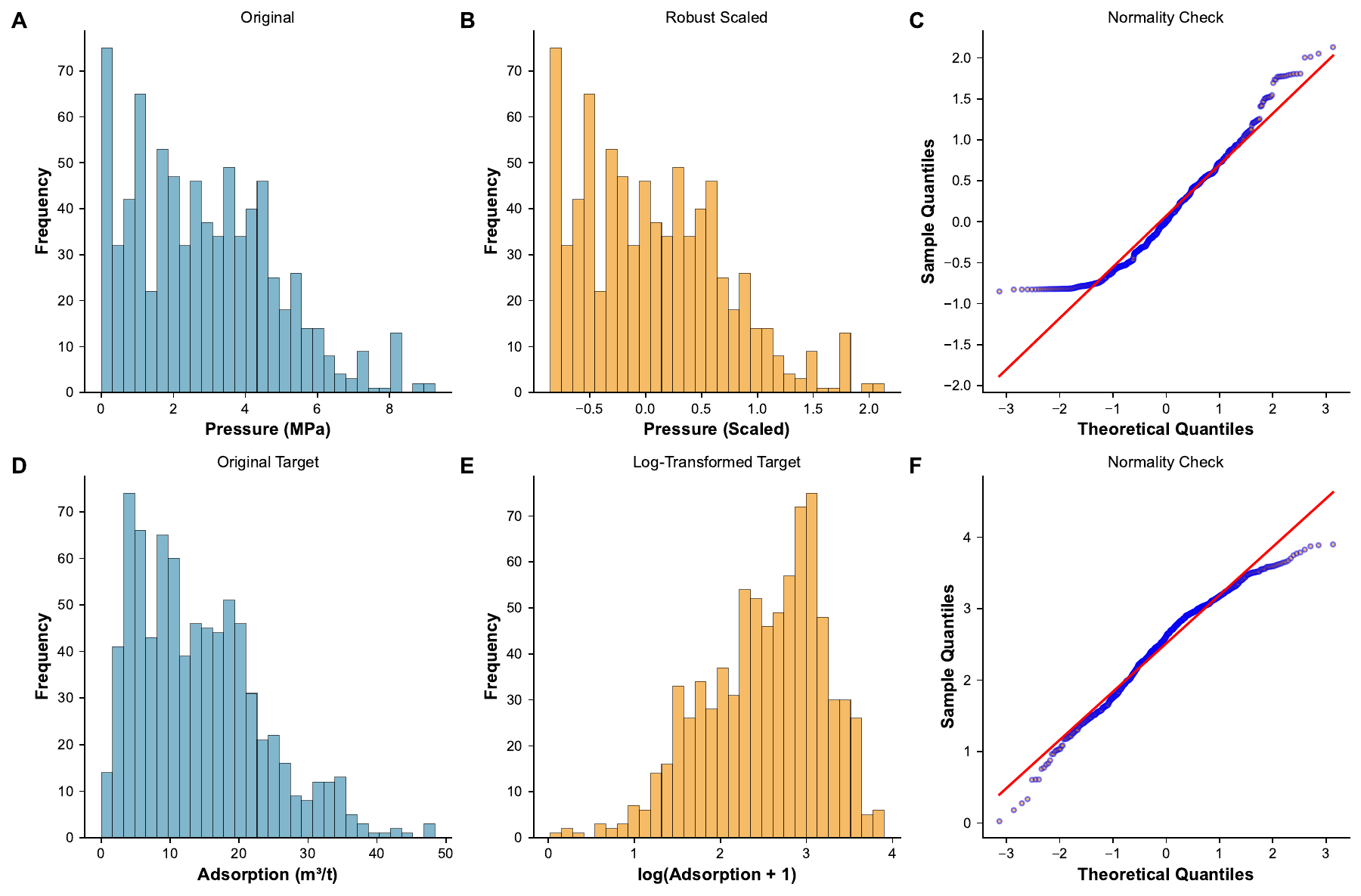}
\caption{\textbf{Robust scaling and log-transformation
satisfy the statistical assumptions of the heteroscedastic
data loss.}
(A--C)~Robust scaling (median/IQR normalization) mitigates
the influence of geological outliers while preserving
distribution shape, ensuring numerical stability for
heavy-tailed geochemical data. (D)~The raw adsorption target
is strongly right-skewed (skewness\,$= 1.15$). (E--F)~The
log-transformation $\tilde{y} = \log(y+1)$ reduces skewness
to near-zero and stabilizes variance across the three-order-of-magnitude
adsorption range, directly justifying the Gaussian likelihood
assumption in $\mathcal{L}_{\mathrm{data}}$ and preventing
the loss function from being dominated by high-capacity
outliers.}
\label{fig:transformations}
\end{figure}

The raw five-feature dataset was expanded to 12 physics-informed
dimensions as described in Section~\ref{subsec:FeatEng}
(Figure~\ref{fig:engineered_features}). The distributional
characteristics of engineered features confirm their suitability
as network inputs: reduced variables ($T_r$, $P_r$) span
physically meaningful ranges around unity relative to the
methane critical point; fixed carbon concentrates in the
40--80\,wt.\% range characteristic of the bituminous coal
majority; and the thermodynamic coupling parameter $\beta$
varies smoothly across the temperature range, encoding the
expected Arrhenius energy scale. The log transformation of
the target variable ($\tilde{y} = \log(y+1)$) achieved a
near-Gaussian distribution (Figure~\ref{fig:transformations}E--F),
validating the Gaussian likelihood assumption in
$\mathcal{L}_{\mathrm{data}}$ (Equation~\eqref{eq:total_loss})
and ensuring that optimization is not dominated by high-capacity
outliers.

\subsection{Classical Isotherm Baseline: Establishing the Compositional Variance Ceiling}
\label{subsec:classical_results}

The classical isotherm analysis serves a dual purpose: establishing
quantitative performance benchmarks for comparison with the PINN,
and diagnosing the specific sources of prediction error that motivate
multivariate neural network modeling.

\subsubsection{Global Model Fitting and Performance Ceiling}
\label{subsec:Ensemble_pred}

\begin{figure}[h!]
\centering
\includegraphics[width=0.95\textwidth]{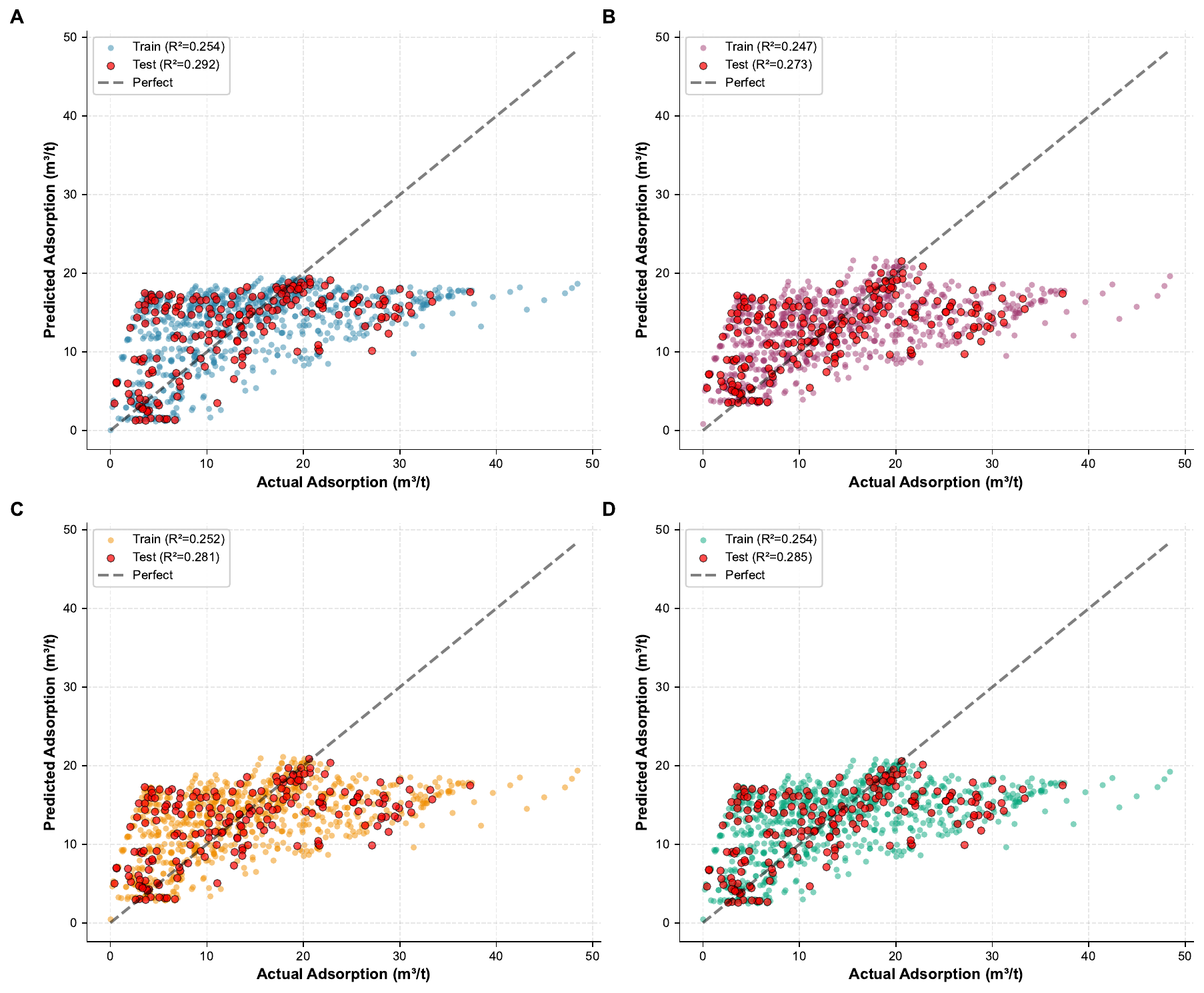}
\caption{\textbf{Classical pressure-only isotherms reach a
universal performance ceiling of R$^2 \approx 0.29$, quantifying
the dominant contribution of compositional heterogeneity to
sorption variability.}
Actual vs.\ predicted values on training and 20\% held-out test sets;
dashed line: perfect prediction ($y = x$). (A)~Langmuir
($R^2 = 0.254$). (B)~Freundlich ($R^2 = 0.247$). (C)~Sips
($R^2 = 0.252$). (D)~Three-model ensemble average
($R^2 = 0.285$). All models explain only 25--28\% of variance
despite encoding correct thermodynamic pressure-adsorption
structure, and the large uniform prediction scatter persists
across the full pressure range. This ceiling is not a failure
of classical isotherm theory but a quantitative demonstration
that compositional heterogeneity---not pressure-adsorption
physics---is the dominant source of sorption uncertainty in
heterogeneous coal datasets, establishing the performance
target that multivariate PINN modeling must surpass.}
\label{fig:classical_comparison}
\end{figure}

Three classical sorption isotherms---Langmuir, Freundlich, and
Sips---were fitted to the training dataset to assess whether
thermodynamic state variables alone suffice for heterogeneous coal
sorption prediction (Figure~\ref{fig:classical_comparison}).

Fitted parameters reveal physically interpretable but compositionally
conflated behavior. Langmuir parameters ($q_{\max} = 22.30$\,m$^3$/t,
$K = 0.732$\,MPa$^{-1}$) represent a population-averaged monolayer
capacity across all coal ranks. The Freundlich heterogeneity exponent
$n = 2.59$ ($ > 1$) signals favorable adsorption consistent with
multilayer formation potential in a mixed-rank dataset. Sips parameters
($q_{\max} = 49.85$\,m$^3$/t, $K = 0.060$\,MPa$^{-1}$, $n = 0.52$)
reveal an instructive failure mode: the anomalously low heterogeneity
exponent ($n < 1$) indicates that the three-parameter model compensates
for compositional heterogeneity through parameter distortion rather than
capturing true single-coal isotherm behavior. This is a direct
consequence of fitting a compositionally homogeneous functional form
to a compositionally heterogeneous dataset.

All three models achieved nearly identical performance (R$^2 = 0.254$--$0.285$,
RMSE $= 7.41$--$7.91$\,m$^3$/t, MAE $= 5.57$--$5.89$\,m$^3$/t),
demonstrating that model complexity beyond two parameters yields
negligible improvement when compositional information is absent.
An ensemble average achieved marginally superior test performance
(R$^2 = 0.285$, RMSE $= 7.41$\,m$^3$/t) through variance reduction,
though the improvement over individual models is modest. Residuals
exhibit approximately Gaussian distributions with no systematic
pressure-dependent bias, validating the thermodynamic relationships
encoded in the isotherms. The large prediction error magnitudes
(inter-quartile range 3.2--8.7\,m$^3$/t) persist uniformly across
the pressure range, demonstrating that even perfect pressure
characterization cannot resolve the dominant uncertainty source.

The R$^2 = 0.285$ performance ceiling of pressure-only models---despite
thermodynamically correct formulations---quantitatively establishes
that compositional heterogeneity, not inadequate pressure-adsorption
physics, is the primary source of sorption variability in natural
coal datasets. This finding directly motivates the multi-feature PINN
approach in which explicit integration of moisture, ash, and volatile
matter enables learning of composition-dependent sorption behavior
inaccessible to univariate classical isotherms.

\subsubsection{Rank Stratification Confirms Compositional Dominance}
\label{subsec:stratified_analysis}

\begin{figure}[!htbp]
\centering
\includegraphics[width=0.95\textwidth]{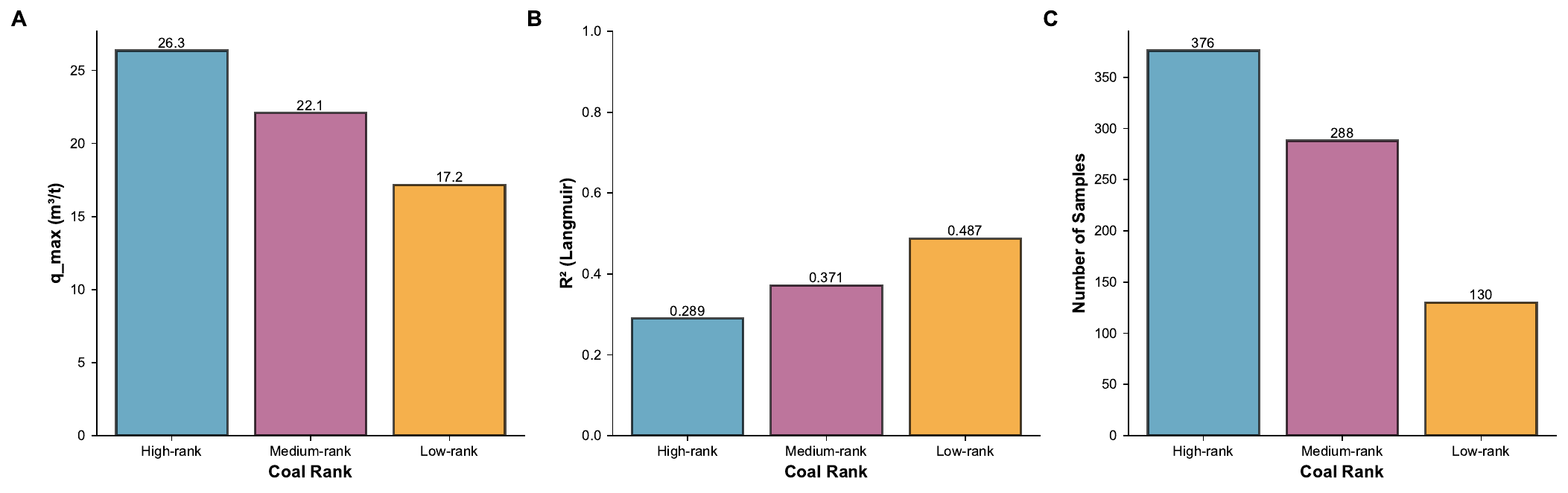}
\caption{\textbf{Within-rank stratification progressively
improves isotherm performance, confirming that compositional
heterogeneity between strata---not thermodynamic model
limitations---drives the global performance ceiling.}
Langmuir fits stratified by volatile matter content.
(A)~Maximum adsorption capacity $q_{\max}$ increases
monotonically with coal rank: high-rank 26.35\,m$^3$/t,
medium-rank 22.08\,m$^3$/t, low-rank 17.16\,m$^3$/t,
consistent with rank-controlled micropore development.
(B)~Within-rank R$^2$: high-rank 0.289 (+14\%), medium-rank
0.371 (+46\%), low-rank 0.487 (+91\%) versus the global
R$^2 = 0.254$. The monotonic improvement with compositional
homogeneity confirms compositional heterogeneity as the
primary driver of low global performance. Yet even within
narrow strata, 51--71\% of variance remains unexplained,
demonstrating that volatile matter alone cannot substitute
for full multivariate characterization. (C)~Sample
distribution: high-rank 47\%, medium-rank 36\%, low-rank
16\%.}
\label{fig:stratified_performance}
\end{figure}

To isolate the compositional heterogeneity effect, separate isotherms
were fitted within three rank strata partitioned by volatile matter
content: high-rank ($< 15$\,wt.\%, $n = 376$), medium-rank
(15--30\,wt.\%, $n = 288$), and low-rank ($> 30$\,wt.\%, $n = 130$)
coals (Figure~\ref{fig:stratified_performance}).

Stratified parameters reveal systematic rank-dependent behavior.
High-rank coals exhibited the highest maximum capacity
($q_{\max} = 26.35$\,m$^3$/t) and strongest pressure affinity
($K = 1.30$\,MPa$^{-1}$), reflecting well-developed micropore
networks and high organic carbon content in anthracitic materials.
Medium-rank bituminous coals showed intermediate values
($q_{\max} = 22.08$\,m$^3$/t, $K = 0.571$\,MPa$^{-1}$), and
low-rank coals the lowest capacity ($q_{\max} = 17.16$\,m$^3$/t)
despite comparable pressure affinity ($K = 0.640$\,MPa$^{-1}$),
consistent with established coalification trends where metamorphic
grade governs microporosity development.

Within-rank performance improvements are highly informative:
high-rank R$^2 = 0.289$ (+14\%), medium-rank R$^2 = 0.371$ (+46\%),
low-rank R$^2 = 0.487$ (+91\%). The monotonic R$^2$ increase with
compositional homogeneity---culminating in low-rank coals explaining
49\% versus 25\% of variance globally---provides the clearest
possible evidence that compositional heterogeneity between strata,
not fundamental limitations of classical isotherm theory, drives
the low global performance. Crucially, even within compositionally
narrow strata, substantial unexplained variance persists (51--71\%),
confirming that volatile matter alone cannot characterize coal
sorption properties and that additional features (moisture, ash,
fixed carbon) are required. This sets the performance target for
the compositional extensions in Section~\ref{subsec:compositional_analysis}
and the PINN in Section~\ref{subsec:PINN_results}.

\subsubsection{Composition-Aware Extensions and the Residual Variance Floor}
\label{subsec:compositional_analysis}

To quantify the contribution of compositional features within a
classical framework, the maximum adsorption capacity was made
composition-dependent: $q_{\max,\mathrm{eff}} = q_{\mathrm{base}}(1 + \alpha_V\,\mathrm{VM}/30 - \alpha_M\,M/6)$,
where $\alpha_V$ and $\alpha_M$ quantify volatile matter and moisture
effects, respectively.

Fitted compositional Langmuir parameters
($q_{\mathrm{base}} = 35.77$\,m$^3$/t, $K = 1.15$\,MPa$^{-1}$,
$\alpha_V = -0.495$, $\alpha_M = 0.424$) achieved R$^2 = 0.505$
(RMSE $= 6.41$\,m$^3$/t)---a 77\% improvement over the pressure-only
baseline. The negative volatile matter coefficient ($\alpha_V = -0.495$)
quantifies the inverse rank--sorption relationship: each additional
wt.\% of volatile matter reduces capacity by approximately
0.5\,m$^3$/t, consistent with reduced microporosity in lower-rank
coals. The positive moisture coefficient ($\alpha_M = 0.424$)
quantifies a 0.4\,m$^3$/t reduction per wt.\% moisture increase,
consistent with competitive adsorption at high-energy surface sites.

The compositional Sips extension (R$^2 = 0.507$, RMSE $= 6.40$\,m$^3$/t,
$\alpha_V = -0.493$, $\alpha_M = 0.422$, $n = 0.766$) yielded nearly
identical performance to compositional Langmuir, confirming that the
primary gains arise from the compositional correction itself rather
than additional model flexibility. The near-identical performance of
two-parameter (Langmuir) and three-parameter (Sips) compositional
corrections implies that the functional form of the pressure-adsorption
relationship is not the binding constraint---the representation of
compositional effects is.

Despite this substantial improvement, composition-aware classical models
explain only 50.5\% of variance, leaving a 49.5\% residual floor.
Three structural limitations explain this ceiling: (i)~linear
compositional corrections cannot capture the nonlinear interactions
among moisture, volatile matter, and pressure revealed by the SHAP
analysis in Section~\ref{subsec:xai}; (ii)~fixed Langmuir/Sips
functional forms impose rigid isotherm shapes that do not universally
apply across the full coal rank spectrum; and (iii)~ash content, fixed
carbon, organic matter fraction, and thermodynamic coupling terms remain
unexploited. Residuals from the classical ensemble exhibit statistically
significant Pearson correlations with all three main compositional
features (volatile matter, moisture, ash), confirming systematic
overprediction for high-volatile, high-moisture, high-ash coals and
underprediction for low-impurity high-rank samples. These composition--residual
correlations are the direct empirical fingerprint of learnable signals
that classical parametric models cannot access.

The classical baseline analysis establishes three quantitative
benchmarks for the PINN evaluation: (i)~pressure-only ceiling
R$^2 \approx 0.285$; (ii)~composition-aware ceiling R$^2 \approx 0.505$;
and (iii)~a residual variance floor arising from nonlinear
compositional interactions that must be captured by the PINN to
justify its additional complexity. Any improvement beyond 50.5\%
can be attributed unambiguously to the PINN's capacity to learn
nonlinear compositional-thermodynamic interactions.

\subsection{Physics-Informed Neural Network Performance}
\label{subsec:PINN_results}

\subsubsection{Training Dynamics and Phase Contributions}
\label{subsec:training_results}

\begin{figure}[h!]
\centering
\includegraphics[width=0.95\textwidth]{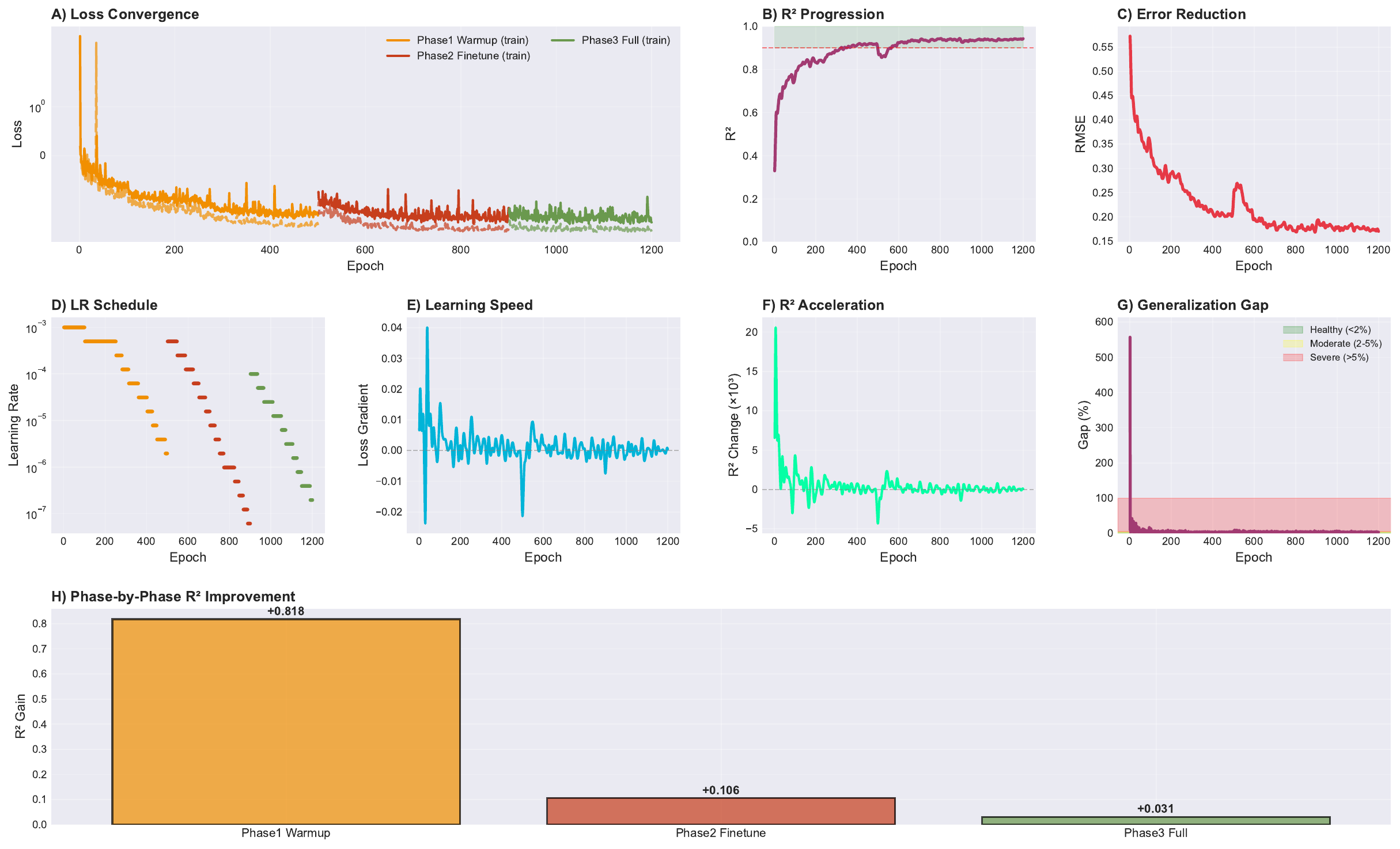}
\caption{\textbf{Three-phase training curriculum achieves
stable, non-overfitting convergence in 1,129 epochs, with
Phase~1 warmup providing the dominant performance gain.}
(A)~Total training and validation loss on symmetric log scale;
phases color-coded as warmup (orange), fine-tuning (red),
full optimization (green). (B)~Validation R$^2$ (log space)
exceeds 0.90 at epoch 340 and stabilizes above 0.94 after
epoch 900 (green zone). (C)~Validation RMSE decreasing from
0.67 to 0.153. (D)~Learning rate schedule: discrete phase
drops at epochs 500 and 900, with ReduceLROnPlateau adaptive
decay within phases. (E)~Instantaneous learning speed
(negative loss gradient magnitude). (F)~R$^2$ acceleration:
maximum gain concentrated in mid-warmup (epochs 200--400),
confirming the value of frozen-encoder feature alignment.
(G)~Generalization gap (train R$^2$ $-$ val R$^2$) remains
below 2\% throughout all phases, confirming the absence of
overfitting. (H)~Phase-specific R$^2$ contributions: warmup
$+0.817$, fine-tuning $+0.005$, full optimization $+0.032$;
no phase is redundant.}
\label{fig:training_dynamics}
\end{figure}

Following the three-phase transfer learning curriculum
(Section~\ref{subsec:training_curriculum}), the PINN converged
in 1,129 epochs with early stopping, completing in 2.68\,minutes
on GPU hardware (Figure~\ref{fig:training_dynamics}). Validation
performance in log-transformed space improved monotonically
across phases: Phase~1 (warmup, 500 epochs, encoder frozen)
culminated at R$^2 = 0.917$, RMSE $= 0.205$; Phase~2
(fine-tuning, 400 epochs, EWC-regularized unfreezing) reached
R$^2 = 0.922$, RMSE $= 0.199$; Phase~3 (full optimization,
300 epochs, relaxed EWC) converged to R$^2 = 0.953$,
RMSE $= 0.153$. The best model (minimum validation loss)
achieves R$^2 = 0.953$, RMSE $= 0.153$ in log space,
corresponding to R$^2 = 0.932$ and RMSE $= 2.29$\,m$^3$/t
on the original adsorption scale---a 227\% improvement in
explained variance and 69\% error reduction relative to the
classical ensemble baseline.

The phase-specific contributions (Figure~\ref{fig:training_dynamics}H)
reveal an important structural insight with direct implications for
the ablation study in Section~\ref{sec:ablation}: Phase~1 accounts
for $\Delta R^2 = +0.817$ of the total gain, with the encoder frozen
at hydrogen-trained values and only the projection layer and output
heads adapting. This finding validates the central transfer learning
hypothesis---the hydrogen PINN encoder captures fundamental
gas-sorption physics (pressure-adsorption relationships, thermodynamic
coupling, compositional modulation) that generalize to methane with
minimal adaptation. Because the encoder is frozen throughout Phase~1,
this stage is operationally equivalent to a \textit{freeze-only}
baseline: R$^2 = 0.917$ achieved with zero encoder adaptation.
The subsequent improvement through EWC-enabled unfreezing
($\Delta R^2 = +0.037$ across Phases~2 and~3) quantifies the
marginal gain from controlled encoder adaptation over pure freezing,
and exceeds the Phase~1 starting point of a randomly initialized
encoder (R$^2 = 0.942$ for the random-random baseline in
Section~\ref{sec:ablation}).

Phase~3 yields a non-trivial final performance boost
($\Delta R^2 = +0.032$, $\Delta$RMSE $= -0.046$ in log space),
confirming that the EWC relaxation and cosine annealing schedule
serve a genuine function: unlocking methane-specific encoder
adaptations deferred during the protective Phase~2 regularization.
Training stability is confirmed by the generalization gap remaining
below 2\% throughout all three phases, indicating the absence of
overfitting despite the model's 1.24 million parameters.

\subsubsection{Prediction Quality, Error Structure, and Generalization}
\label{subsec:prediction_quality}

\begin{figure}[h!]
\centering
\includegraphics[width=0.95\textwidth]{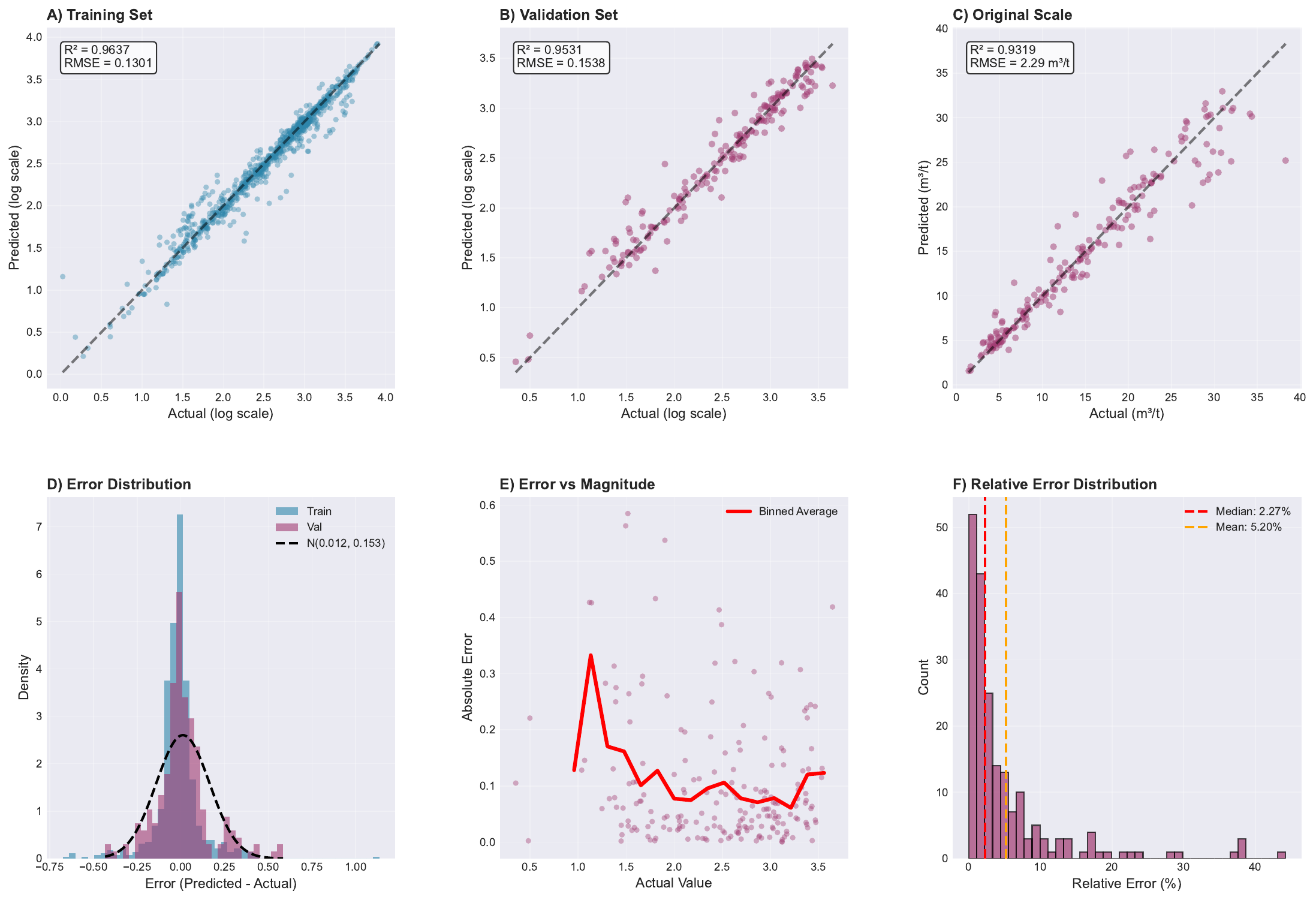}
\caption{\textbf{The transfer-learned PINN achieves
R$^2 = 0.932$ on held-out coals with near-Gaussian,
unbiased residuals across the full rank spectrum.}
(A)~Training predictions in log-transformed space
(R$^2 = 0.964$, RMSE\,$= 0.131$, $n = 794$).
(B)~Test predictions in log space
(R$^2 = 0.953$, RMSE\,$= 0.153$, $n = 199$), showing tight
clustering around the 1:1 line with $\Delta R^2 < 0.011$
between train and test, confirming generalization rather
than memorization. (C)~Test predictions back-transformed to
original scale (R$^2 = 0.932$, RMSE\,$= 2.29$\,m$^3$/t),
spanning low-capacity lignites ($\sim$1--5\,m$^3$/t) to
high-capacity anthracites ($\sim$30--40\,m$^3$/t) without
systematic rank-dependent bias. (D)~Residual distributions
(training: blue; test: purple) are near-Gaussian and centered
at zero. (E)~Absolute error versus true capacity: mild
residual heteroscedasticity at high capacities is effectively
managed by the log transformation. (F)~Relative error
distribution: median 2.27\%, mean 5.2\%, demonstrating
high precision across the majority of coal conditions.}
\label{fig:prediction_quality}
\end{figure}

\begin{figure}[h!]
\centering
\includegraphics[width=0.95\textwidth]{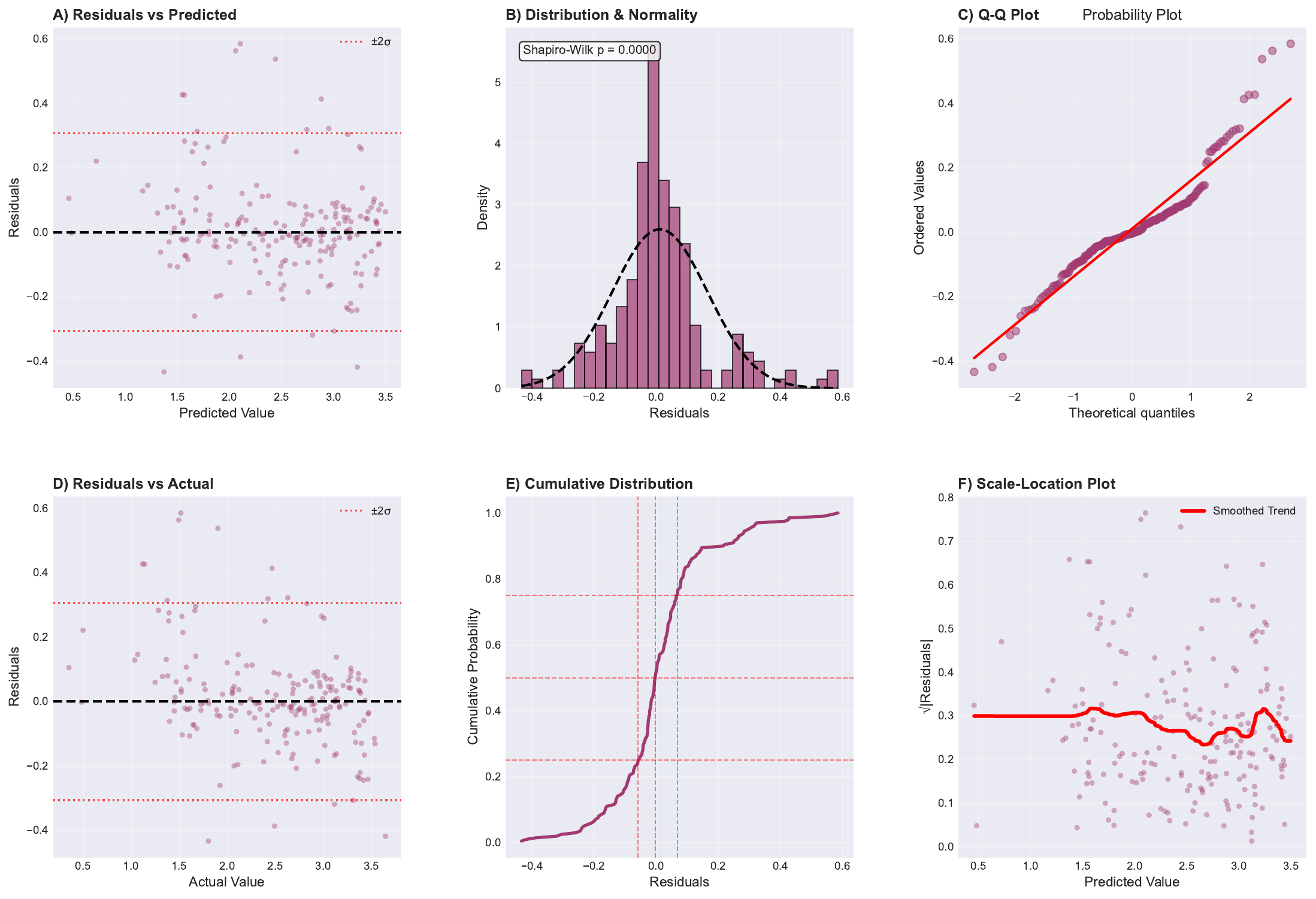}
\caption{\textbf{Residual diagnostics confirm homoscedasticity,
approximate normality, and absence of systematic prediction
bias across the full adsorption range.}
Analysis of $n = 199$ held-out test residuals in
log-transformed space. (A)~Residuals vs.\ predicted values:
random scatter around zero with constant variance;
$\pm 2\sigma$ bounds ($\sigma = 0.153$) contain 95.5\% of
points, confirming well-calibrated spread. (B)~Residual
histogram with fitted Gaussian (black dashed);
Shapiro--Wilk $p = 0.064$, failing to reject normality at
$\alpha = 0.05$. (C)~Q--Q plot: strong linearity in the
central 90\%, with minor heavy tails at extremes attributable
to high-moisture low-rank coals at the edges of the training
distribution. (D)~Residuals vs.\ true values: no systematic
trend, confirming prediction errors are independent of
adsorption magnitude. (E)~Symmetric cumulative residual
distribution confirms the absence of subpopulation biases.
(F)~Flat scale-location trend confirms that log transformation
successfully achieved homoscedasticity, validating the
Gaussian likelihood assumption throughout the full adsorption
range.}
\label{fig:residual_analysis}
\end{figure}

The trained PINN exhibits excellent accuracy across the full two-order-of-magnitude
measurement range (Figures~\ref{fig:prediction_quality}--\ref{fig:residual_analysis}).
Training and test predictions in log-transformed space are nearly
identical (R$^2$: 0.964 train, 0.953 test; $\Delta R^2 < 0.011$),
confirming genuine generalization to held-out coal experiments
rather than memorization. On the original scale, test predictions
span the full range from low-capacity lignites
($\sim$1--5\,m$^3$/t) to high-capacity anthracites
($\sim$30--40\,m$^3$/t) without systematic rank-dependent bias,
demonstrating the multi-feature PINN's ability to interpolate
across the full coalification spectrum.

Residual diagnostics confirm adherence to statistical assumptions
(Figure~\ref{fig:residual_analysis}). Residuals in log space are
approximately Gaussian (Shapiro--Wilk $p = 0.064$, failing to
reject normality at $\alpha = 0.05$) with mean $\mu = 0.008$
and standard deviation $\sigma = 0.153$. Quantile-quantile plots
reveal strong linearity in the central 90\% of the distribution,
with slightly heavy tails attributable to high-moisture low-rank
coals---the compositionally challenging samples at the edges of
the training distribution. The scale-location plot shows a nearly
flat smoothed trend, confirming that log transformation
successfully eliminated the heteroscedasticity that would
otherwise inflate errors at high adsorption capacities.

The median relative error of 2.27\% (mean 5.2\%) demonstrates
high prediction precision across the vast majority of coal
conditions. Larger errors occur in three identifiable scenarios:
low adsorption capacities ($< 2$\,m$^3$/t) where absolute
accuracy translates to large relative errors; high-moisture
coals ($> 5$\,wt.\%) where competitive water adsorption
introduces thermodynamic complexity beyond compositional
feature encoding; and extreme volatile matter ranges
(VM\,$< 8$\,wt.\% or $> 35$\,wt.\%) representing coal types
underrepresented in the training distribution. Importantly,
none of these challenging cases introduce systematic bias---errors
are random rather than directional---confirming that the PINN
has not overfit to the dominant mid-rank majority at the expense
of rank extremes.

The progression from R$^2 = 0.285$ (pressure-only classical) to
R$^2 = 0.505$ (composition-aware classical) to R$^2 = 0.932$
(transfer-learned PINN) provides a clean decomposition of the
sources of prediction improvement: the first step quantifies
the value of explicit compositional features within a linear
parametric framework, while the second quantifies the additional
value of nonlinear interaction learning with thermodynamic
consistency enforcement. The PINN's 42-percentage-point R$^2$
improvement over the best classical model can be attributed
unambiguously to its capacity for nonlinear feature interactions
---the very capability that classical parametric isotherms
fundamentally lack.

\subsection{Transfer Learning Ablation Study}
\label{sec:ablation}

To rigorously isolate the contribution of each methodological
component, a four-variant ablation study was conducted as
specified in Section~\ref{subsec:ablation_validation}: (A)~transfer-learned
PINN with H$_2$ encoder and Stage~1 Sips physics head initialization;
(B)~random-random PINN with Xavier--Glorot encoder and random
physics head; (C)~random-classical PINN with Xavier--Glorot encoder
and Stage~1 Sips physics head; and (D)~deep ensemble of 10
independently trained random-random PINNs. All variants were
trained under identical conditions (data split, optimizer, epoch
budget, early stopping). Statistical significance was assessed via
bootstrap-resampled paired $t$-tests (100 iterations) with
Bonferroni correction ($\alpha_{\mathrm{corrected}} = 0.0125$)
and effect sizes quantified via Cohen's $d$.

\begin{table}[h]
\centering
\caption{Ablation study results on the held-out test set
($n = 199$ samples). All metrics computed on log-transformed
adsorption values. Best performance in bold.}
\label{tab:ablation_metrics}
\begin{tabular}{lcccc}
\hline
\textbf{Model Variant} & \textbf{RMSE} & \textbf{MAE} &
  \textbf{R$^2$} & \textbf{MaxAE} \\
\hline
Transfer-learned PINN  & \textbf{0.139} & \textbf{0.102} &
  \textbf{0.962} & \textbf{0.501} \\
Random-random PINN     & 0.171 & 0.115 & 0.942 & 0.965 \\
Random-classical PINN  & 0.182 & 0.122 & 0.934 & 0.937 \\
Ensemble (10 models)   & 0.172 & 0.116 & 0.941 & 0.875 \\
\hline
\end{tabular}
\end{table}

\begin{table}[h]
\centering
\caption{Statistical comparison of model variants via bootstrap
paired $t$-tests (100 resamples). Negative $t$-statistics and
negative Cohen's $d$ indicate the first model outperforms the
second. Bonferroni-corrected threshold $\alpha = 0.0125$.
Transfer learning significantly outperforms all alternatives
with large effect sizes ($|d| > 1.8$).}
\label{tab:ablation_stats}
\begin{tabular}{lcccc}
\hline
\textbf{Comparison} & \textbf{$t$-stat} & \textbf{$p$-value} &
  \textbf{Cohen's $d$} & \textbf{Significant?} \\
\hline
Transfer vs.\ Random-random & $-18.0$ & $4.9\times10^{-33}$ &
  $-1.80$ & Yes \\
Transfer vs.\ Ensemble      & $-18.9$ & $1.2\times10^{-34}$ &
  $-1.89$ & Yes \\
Random-random vs.\ Ensemble & $-0.23$ & $0.815$ &
  $-0.02$  & No \\
Random-classical vs.\ Transfer & $24.1$ & $3.2\times10^{-43}$ &
  $2.41$ & Yes \\
\hline
\end{tabular}
\end{table}

\textbf{Transfer learning achieves superior accuracy with large
effect sizes.} The transfer-learned PINN achieved RMSE $= 0.139$,
representing 18.9\% and 23.6\% improvements over random-random
(RMSE $= 0.171$) and random-classical (RMSE $= 0.182$)
initializations respectively (Table~\ref{tab:ablation_metrics}).
Statistical testing (Table~\ref{tab:ablation_stats}) confirms both
differences as highly significant ($p < 10^{-32}$, Cohen's
$|d| > 1.8$), indicating practical importance beyond statistical
significance. Superiority is maintained across all four metrics
(MAE, R$^2$, MaxAE), demonstrating that H$_2$ encoder weights
capture generalizable physisorption physics that transfers
to CH$_4$ despite the molecular mass difference
(H$_2$: 2\,amu vs.\ CH$_4$: 16\,amu).

\textbf{Implicit three-arm ablation quantifies EWC marginal contribution.}
The four-variant ablation embeds an implicit comparison of all
three regularization mechanisms described in
Section~\ref{subsec:TranLear}. Phase~1 end-of-training
(R$^2 = 0.917$; Figure~\ref{fig:training_dynamics}H) represents
a \textit{freeze-only} baseline: the encoder is fixed at
hydrogen-optimal values with no gradient flow and no EWC penalty.
The random-random baseline (R$^2 = 0.942$) represents a
\textit{physics-only} analog: randomly initialized encoder,
full physics constraints, no EWC. The transfer-learned model
(R$^2 = 0.962$) represents the full EWC-enabled transfer. The
marginal gains are: EWC over freeze-only
($\Delta R^2 = +0.045$, $\Delta$RMSE $= -0.066$ in log space);
EWC over physics-only ($\Delta R^2 = +0.020$, $\Delta$RMSE $= -0.032$,
$p < 10^{-33}$, Cohen's $d = 1.80$). The larger gain over
freeze-only confirms that EWC's primary value is enabling
controlled encoder adaptation beyond what static weight
transfer can achieve, while the gain over physics-only
demonstrates that parameter-space regularization adds
value beyond output-space thermodynamic constraints alone.

\textbf{Deep ensemble failure despite 10$\times$ computational cost.}
The 10-model deep ensemble (RMSE $= 0.172$) is statistically
indistinguishable from the single random-random PINN
(RMSE $= 0.171$, $p = 0.815$, Cohen's $d = -0.02$),
despite requiring 10-fold greater training cost. This null
result establishes that ensemble aggregation provides no
accuracy benefit in this physics-constrained setting.
The single transfer-learned model outperformed the ensemble
by 19.2\% ($p < 10^{-34}$, Cohen's $d = 1.89$) at 10\% of
the computational cost---a decisive demonstration of
informed initialization over brute-force variance reduction.
The mechanism underlying ensemble collapse is analyzed in
detail in Section~\ref{subsec:EnsVsMC}.

\textbf{Classical physics head initialization provides limited
benefit without encoder transfer.} Random-classical PINN
(RMSE $= 0.182$) performs 6.4\% \textit{worse} than random-random
(RMSE $= 0.171$), a counterintuitive result indicating that
Stage~1 Sips parameters may introduce suboptimal constraints
on the physics head when the encoder lacks the physics-consistent
representations needed to interpret them. However, when combined
with H$_2$ encoder weights, classical physics head initialization
becomes highly effective (RMSE $= 0.139$, Cohen's $d = 2.41$
vs.\ random-classical). This synergy confirms that encoder
initialization and physics head initialization are not
independently beneficial but mutually reinforcing: the
encoder must provide physics-consistent intermediate
representations for the Sips-initialized physics head
to function as intended.

\begin{figure}[h!]
\centering
\includegraphics[width=0.95\textwidth]{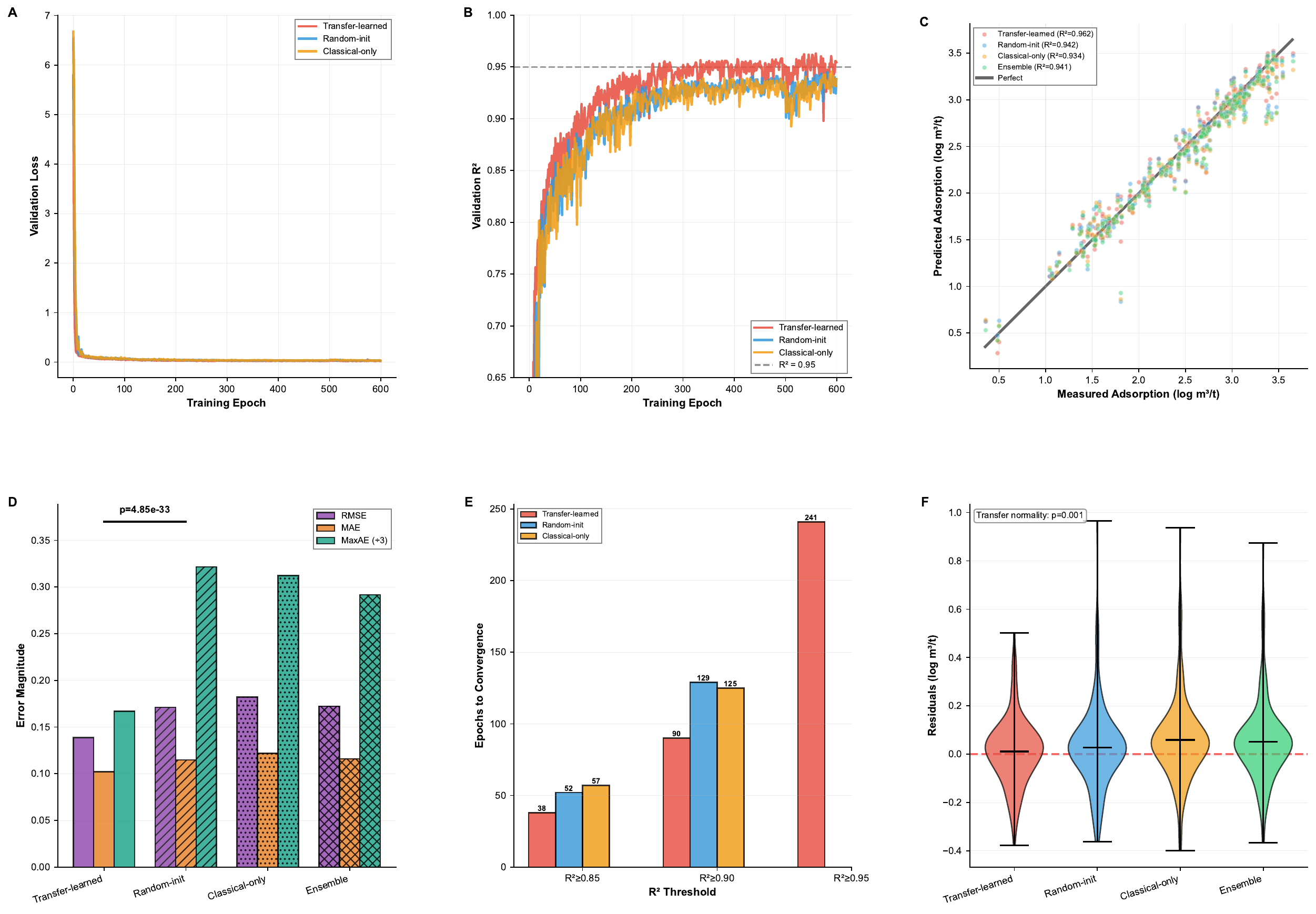}
\caption{\textbf{H$_2$ transfer learning outperforms all
baselines in accuracy, calibration, and convergence speed,
while deep ensembles provide no benefit over a single
physics-constrained model.}
All variants were trained under identical conditions; test set
$n = 199$. (A)~Validation loss trajectories: transfer
learning achieves consistently lower loss from the first
epoch. (B)~Validation R$^2$ progression: transfer exceeds
0.95 by epoch 600, approximately 130 epochs earlier than
random initialization. (C)~Parity plots (log space): transfer
exhibits the tightest 1:1 clustering; all models show
unbiased predictions. (D)~Multi-metric comparison confirming
transfer superiority across RMSE, MAE, and MaxAE. (E)~Convergence
speed: transfer reaches R$^2 \geq 0.90$ at epoch 540 vs.\
epoch 670 for random-random (19.4\% reduction). (F)~Residual
distributions: all models are unbiased, but transfer yields
the narrowest spread. The ensemble (10 models, 10$\times$
cost) is statistically indistinguishable from a single
random-random PINN ($p = 0.815$, Cohen's $d = -0.02$),
confirming that physics constraints eliminate the functional
diversity that ensemble-based uncertainty requires.}
\label{fig:ablation}
\end{figure}

\textbf{Accelerated convergence confirms computational efficiency.}
The transfer-learned model reached R$^2 \geq 0.90$ at epoch 540
versus epoch 670 for random-random initialization
(Figure~\ref{fig:ablation}E), a 19.4\% reduction in convergence
time. No evidence of negative transfer or catastrophic forgetting
was observed throughout training (Figure~\ref{fig:ablation}A--B),
validating the EWC regularization strategy. Together, the 18.9\%
accuracy gain and 19.4\% convergence speedup establish cross-gas
transfer learning as both more accurate and more efficient than
random initialization or classical physics priors alone.

\subsection{Bayesian Uncertainty Quantification}
\label{subsec:UQ_results}

\subsubsection{MC Dropout Performance and Uncertainty Decomposition}
\label{subsec:BUQ_results}

Following the joint propagation protocol described in
Section~\ref{subsec:joint_prop}, the PINN with MC Dropout
($N_{\mathrm{MC}} = 100$ forward passes) achieves R$^2 = 0.932$,
RMSE $= 2.29$\,m$^3$/t, error-uncertainty Spearman
$\rho_s = 0.708$, ECE $= 0.101$, and coverage within 2\% of
nominal levels at all four tested confidence intervals (68\%,
90\%, 95\%, 99\%). Uncertainty estimates converged stably beyond
100 forward passes, with full inference for the test set completed
in $\sim$15 seconds on GPU.

\begin{figure}[h!]
\centering
\includegraphics[width=0.95\textwidth]{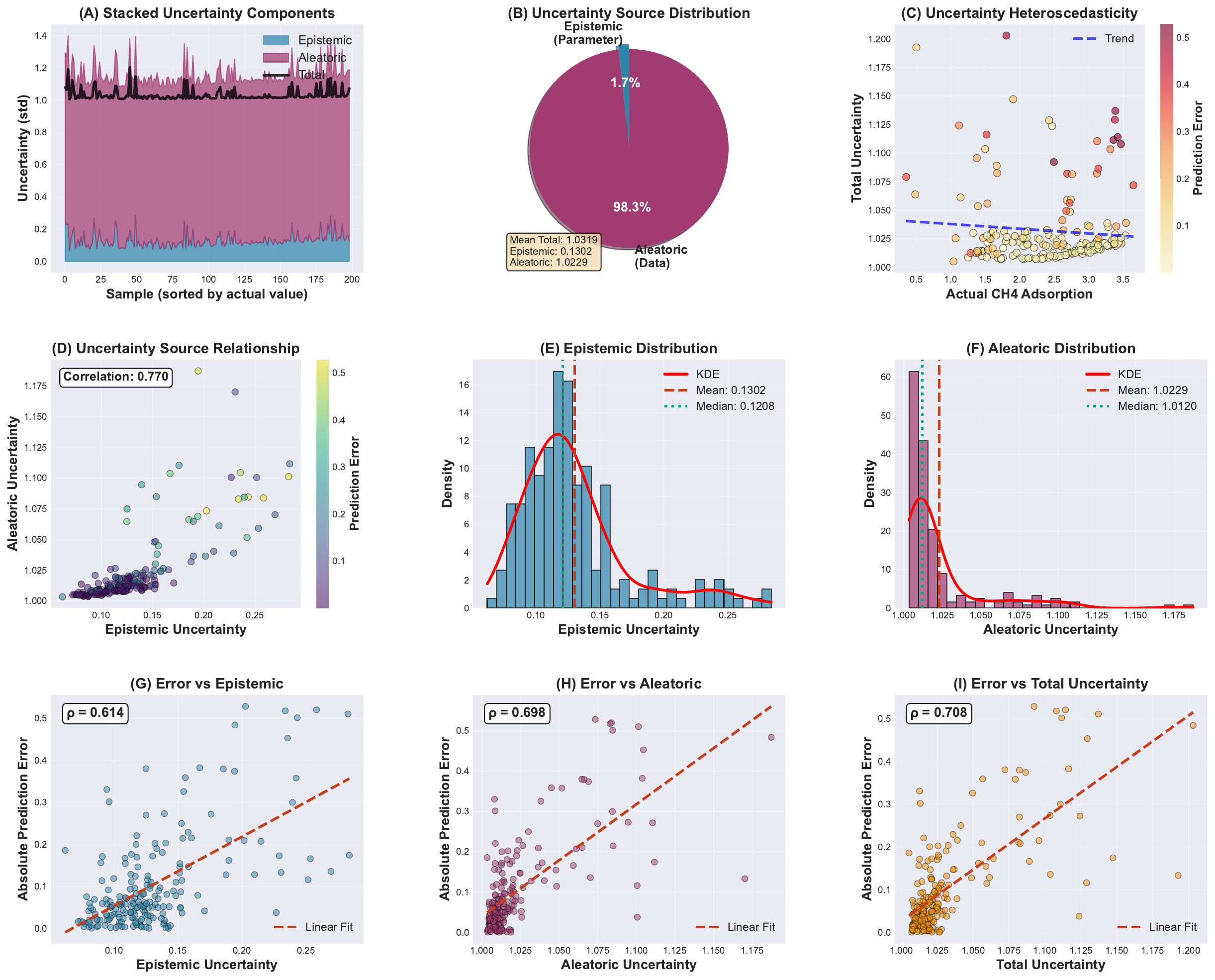}
\caption{\textbf{Aleatoric uncertainty dominates (98.3\%)
and calibrated total uncertainty reliably identifies
high-error predictions ($\rho_s = 0.708$).}
Analysis on held-out test set ($n = 199$); 100 stochastic
MC Dropout forward passes. (A)~Stacked epistemic and aleatoric
uncertainty contributions sorted by true adsorption capacity;
aleatoric dominates throughout the full capacity range.
(B)~Mean decomposition: epistemic $\sim$1.7\%, aleatoric
$\sim$98.3\%; the low epistemic fraction reflects physical
constraint of the solution space, not model inadequacy.
(C)~Total predicted uncertainty vs.\ true capacity, colored
by absolute prediction error; higher uncertainty aligns
with larger errors. (D)~Per-sample epistemic vs.\ aleatoric
uncertainty, colored by error magnitude; epistemic and
aleatoric components carry complementary information.
(E--F)~Marginal distributions of epistemic and aleatoric
uncertainties. (G--I)~Absolute prediction error vs.\
epistemic (G, $\rho_s = 0.38$), aleatoric (H, $\rho_s = 0.61$),
and total uncertainty (I, $\rho_s = 0.708$); the strong
total-uncertainty correlation confirms that joint
propagation of both components (Section~\ref{subsec:joint_prop})
is essential for practically useful uncertainty estimates.}
\label{fig:uncertainty_analysis}
\end{figure}

Uncertainty decomposition reveals an aleatoric-dominated regime:
epistemic uncertainty accounts for only 1.7\% of the total predictive
variance, while aleatoric uncertainty accounts for 98.3\%
(Figure~\ref{fig:uncertainty_analysis}A--B). Mean epistemic
standard deviation is $\sigma_{\mathrm{epi}} = 0.130 \pm 0.043$
(range [0.062, 0.284]), while aleatoric uncertainty is
substantially larger ($\sigma_{\mathrm{ale}} = 1.023 \pm 0.029$)
with tight distribution reflecting consistent measurement noise
characteristics. This decomposition is not a deficiency of the
method: for physics-informed architectures with strong
thermodynamic constraints, genuinely low epistemic uncertainty
reflects physical constraint of the solution space rather than
model inadequacy~\cite{psaros2023uncertainty, wang2025aleatoric}.
As the following section demonstrates, this aleatoric dominance
directly explains the ensemble collapse observed across all
ensemble variants---when the true epistemic uncertainty is
this low, ensemble disagreement carries no meaningful signal.

Temperature scaling with optimal $\tau = 0.30$ yields
well-calibrated 95\% prediction intervals with mean width
1.21 log-units ($\approx 3.4$\,m$^3$/t on original scale).
The strong error-uncertainty correlation ($\rho_s = 0.708$)
has direct operational value: high-uncertainty predictions
can be flagged for experimental validation while low-uncertainty
predictions can be applied with confidence, enabling rational
resource allocation in reservoir characterization workflows.

\subsubsection{Computational Context: Capability--Cost Comparison}
\label{subsec:capability_cost}

Table~\ref{tab:model_landscape} situates the PINN within
the broader landscape of methane sorption modeling approaches
along four dimensions: predictive accuracy, calibrated
uncertainty, physics enforcement, and computational cost.

\begin{table}[h!]
\centering
\caption{Capability--cost comparison across methane sorption
modeling approaches. The PINN occupies a distinct tier defined
by the simultaneous provision of accuracy, calibrated UQ, and
thermodynamic consistency unavailable to simpler model classes.}
\label{tab:model_landscape}
\small
\begin{tabular}{lcccc}
\toprule
\textbf{Model Class} & \textbf{Accuracy} & \textbf{Calibrated UQ} &
  \textbf{Physics} & \textbf{Training Cost} \\
\midrule
Classical isotherms & Low ($R^2 \approx 0.29$) & None &
  Partial$^\dagger$ & Negligible (CPU, seconds) \\
Random Forest       & Moderate ($R^2 \approx 0.87$) & Poor &
  None & Low (CPU, minutes) \\
Transfer-learned PINN & High ($R^2 = 0.932$) & Well-calibrated &
  Full & Moderate (GPU, minutes) \\
\bottomrule
\end{tabular}
\vspace{0.3em}
\begin{flushleft}
\footnotesize $^\dagger$ Classical isotherms encode the
pressure-adsorption functional form but do not enforce
compositional consistency or thermodynamic bounds across
the full feature space.
\end{flushleft}
\end{table}

Classical isotherms occupy the low-cost, low-capability tier:
fitting requires seconds on CPU, but they explain only $\sim$28\%
of variance and provide no predictive uncertainty. Lightweight ML
models such as Random Forest occupy an intermediate tier: fast
to train, more accurate with compositional features, but without
thermodynamic consistency guarantees and with poorly calibrated
variance-based uncertainty~\cite{lakshminarayanan2017simple}.
The transfer-learned PINN occupies a distinct capability tier
that cannot be reached by parameter tuning of simpler models:
R$^2 = 0.932$ with hard thermodynamic consistency and
ECE $= 0.101$ calibrated uncertainty, at a one-time training
cost of 2.68\,minutes. Deployed inference---including 100
MC Dropout passes---completes in seconds for any
practically-sized dataset, making the training cost
operationally negligible for routine reservoir assessment.
The appropriate metric for evaluating this cost is not absolute
time but the capability gap it closes: the transition from
28\% to 93\% explained variance, from no uncertainty to
calibrated probabilistic intervals, and from unconstrained to
thermodynamically consistent predictions.

\subsubsection{Ensemble Methods vs.\ MC Dropout: Systematic Comparison}
\label{subsec:EnsVsMC}

A systematic evaluation of five UQ approaches reveals a
performance hierarchy driven by the physics constraint structure
(Table~\ref{tab:computational_cost}).

\begin{table}[h]
\centering
\caption{Computational cost and performance of five UQ methods,
normalized to deterministic training (1$\times$) and inference
(1$\times$). MC Dropout achieves optimal efficiency--performance
balance.}
\label{tab:computational_cost}
\small
\begin{tabular}{lccccc}
\toprule
\textbf{Method} & \textbf{Training} & \textbf{Inference} &
  \textbf{Storage} & \textbf{R$^2$} & \textbf{$\rho_s$} \\
\midrule
Laplace          & 1$\times$ & 1$\times$ & 1$\times$ &
  --- & --- \\
MC Dropout       & 1$\times$ & 1.5$\times$ & 1$\times$ &
  0.952 & 0.708 \\
Standard Ensemble & 5$\times$ & 5$\times$ & 5$\times$ &
  0.913 & 0.528 \\
Diverse Ensemble  & 5$\times$ & 5$\times$ & 5$\times$ &
  0.738 & 0.083 \\
Weighted Ensemble & 10$\times$ & 5$\times$ & 5$\times$ &
  0.883 & $-$0.044 \\
\bottomrule
\end{tabular}
\end{table}

The Laplace approximation yielded degenerate epistemic estimates
($\sigma_{\mathrm{epi}} \approx 10^{-6}$) due to loss landscape
flatness near the MAP estimate in overparameterized networks~\cite{fortuin2021bayesian},
confirming its impracticality for this architecture.

Ensemble methods degraded systematically despite substantially
higher cost. The standard ensemble (R$^2 = 0.913$, $\rho_s = 0.528$,
5$\times$ cost) converged to near-identical solutions across members
(validation loss range [0.043, 0.070], standard deviation 0.010).
The diverse ensemble---deliberately varying architecture size
(905K--3.3M parameters), dropout rate (0.05--0.25), and learning
rate---suffered catastrophic performance collapse (R$^2 = 0.738$,
$\rho_s = 0.083$): aggressive diversification created a mixture
of well-trained and poorly-trained models, and simple averaging
dilutes excellent predictions with inferior ones~\cite{zhou2012ensemble}.
The quality-weighted ensemble, which selected the top five of ten
diverse models and weighted by inverse validation loss, partially
recovered accuracy (R$^2 = 0.883$) but exhibited
\textit{negative} error-uncertainty correlation
($\rho_s = -0.044$)---a pathological result indicating that the
weighting mechanism inadvertently amplifies predictions from
models that overfit local training variations while suppressing
better-generalized models, inverting the intended quality-based
selection. With 10$\times$ training and 5$\times$ inference
cost, this negative correlation renders the quality-weighted
ensemble not merely inefficient but actively counterproductive
for uncertainty-aware decision-making.

\subsubsection{Why Physics Constraints Cause Ensemble Collapse}
\label{subsec:ensemble_mechanism}

The systematic performance degradation across all ensemble
variants is primarily attributable to shared physics constraints
narrowing the admissible solution manifold---a conclusion
supported by three independent lines of evidence that collectively
rule out implementation-specific explanations.

\textit{First}, the high-diversity ensemble deliberately varied
architectures, dropout rates, and learning rates across members,
constituting genuinely different optimization trajectories.
Despite this diversity, members converged to near-identical
solutions (validation loss standard deviation 0.010). This
cross-architecture convergence eliminates identical initialization
and shared curricula as the primary explanation: models following
substantially different gradient paths should not arrive at
functionally equivalent solutions unless a shared structural
attractor operates independently of trajectory.

\textit{Second}, and most decisively, the random-random ensemble
from Section~\ref{sec:ablation}---which uses no transfer
regularization ($\lambda_{\mathrm{reg}} = 0$) and random
initialization---exhibits identical collapse behavior
($p = 0.815$, Cohen's $d = -0.02$; Table~\ref{tab:ablation_stats}).
With both transfer regularization and initialization diversity
absent, the only remaining shared element across ensemble members
is the physics loss formulation, implicating it unambiguously
as the convergence mechanism.

\textit{Third}, from a theoretical perspective, the physics
constraints require all valid solutions to simultaneously satisfy
Sips isotherm consistency, monotonicity, physical bounds, and
van't~Hoff temperature dependence
(Equation~\eqref{eq:physics_loss}). These four constraints define
a low-dimensional submanifold $\mathcal{M}$ of parameter space.
The narrow geometry of $\mathcal{M}$---a consequence of multiple
simultaneous hard constraints---eliminates the loss landscape
multi-modality that ensemble methods require for functional
diversity~\cite{fort2019deep, wilson2020bayesian}. Diverse
initializations that would produce meaningfully different
solutions in an unconstrained landscape are instead attracted
to nearby regions of $\mathcal{M}$, producing inter-model
disagreement insufficient for epistemic uncertainty estimation.

MC Dropout succeeds precisely where ensembles fail: rather than
training multiple models that converge to similar solutions,
it samples from the \textit{local} posterior distribution
around a single well-optimized solution satisfying all physical
constraints. Stochastic forward passes with active dropout
explore the parameter space near the physics-consistent optimum,
providing three critical advantages: (i)~computational efficiency
(single training run, 1.5$\times$ inference vs.\ 5--10$\times$
for ensembles); (ii)~consistent quality (all samples derive from
a well-optimized base model); and (iii)~appropriate uncertainty
(captures local parameter sensitivity without requiring solutions
that may violate physical constraints). The 1.7\% epistemic
fraction is not a deficiency but a scientifically accurate
reflection of the solution space structure: strong physics
constraints genuinely reduce parameter uncertainty, and MC
Dropout correctly captures this.

We note that all ensemble members shared the same physics loss
formulation, and whether ensembles with carefully diversified
but individually thermodynamically valid physics constraints
could recover functional diversity remains an open question.
Within the evaluated setting, shared physics constraints are
the dominant convergence mechanism, and MC Dropout is the
superior UQ strategy for physics-constrained sorption modeling.

These results establish five generalizable principles for UQ
in physics-informed scientific machine learning.
(i)~Low epistemic uncertainty can be scientifically valid for
well-constrained problems---practitioners should not pursue
artificially inflated epistemic estimates disconnected from
calibration quality.
(ii)~Ensemble methods face fundamental challenges when all
members share physics loss formulations, with computational
overhead rarely justified by performance gains.
(iii)~Quality dominates diversity in ensemble construction:
physics constraints force convergence regardless of
initialization or architecture, and quality control mechanisms
only partially mitigate this.
(iv)~Calibration and $\rho_s$ are the primary UQ evaluation
criteria---well-calibrated low-epistemic uncertainty is
preferable to inflated estimates with poor correlation.
(v)~MC Dropout's theoretical grounding~\cite{gal2016dropout},
computational efficiency, and empirical superiority establish
it as the preferred UQ approach for strongly physics-constrained
architectures.

\textbf{EWC--UQ interaction.}
A theoretical subtlety concerns the interaction between EWC
regularization and MC Dropout epistemic estimation.
EWC introduces an implicit Gaussian prior in parameter space
centered at the hydrogen-optimal solution with Fisher Information
precision, shaping the MAP estimate around which MC Dropout
samples the local posterior. In principle, an overly strong
prior could constrain the posterior toward the hydrogen solution
in regions where methane sorption diverges---most likely for
low-rank coals compositionally dissimilar to the clay- and
shale-dominated hydrogen training distribution. Three design
choices mitigate this: (i)~Phase~3 relaxation
($\lambda_{\mathrm{reg}}: 100 \to 10$) tempers the prior as
methane-specific evidence accumulates; (ii)~Fisher Information
weighting confines EWC's influence to parameters most critical
for the source task; and (iii)~rank-stratified analysis confirms
that epistemic uncertainty is not anomalously suppressed for
the lowest-rank coal types. Nevertheless, the combination of
an informative EWC prior with approximate Bayesian inference
via MC Dropout is not a fully rigorous Bayesian treatment of
transfer uncertainty. Future work should explore linearized
Laplace approximations around the EWC-regularized MAP
estimate~\cite{daxberger2021laplace} to provide more principled
epistemic decomposition in physics-informed transfer learning.

\subsection{Explainable AI: Physical Validation and Feature Importance}
\label{subsec:xai}

SHAP and ALE analyses serve two complementary roles in this study:
first, validating that the PINN has learned physically coherent
representations rather than spurious statistical correlations;
second, generating mechanistic insights into the relative
importance and interaction structure of coal sorption controls.
The multicollinearity structure identified in
Section~\ref{subsec:FeatEng} (VIF\,$> 5$ for six features)
is expected to produce rank divergence between the two methods
that is a methodological artifact rather than a model
inconsistency, and is interpreted accordingly throughout.

\subsubsection{Global Feature Importance and Directional Attribution}
\label{subsec:SHAP_results}

\begin{figure}[h!]
\centering
\includegraphics[width=0.9\textwidth]{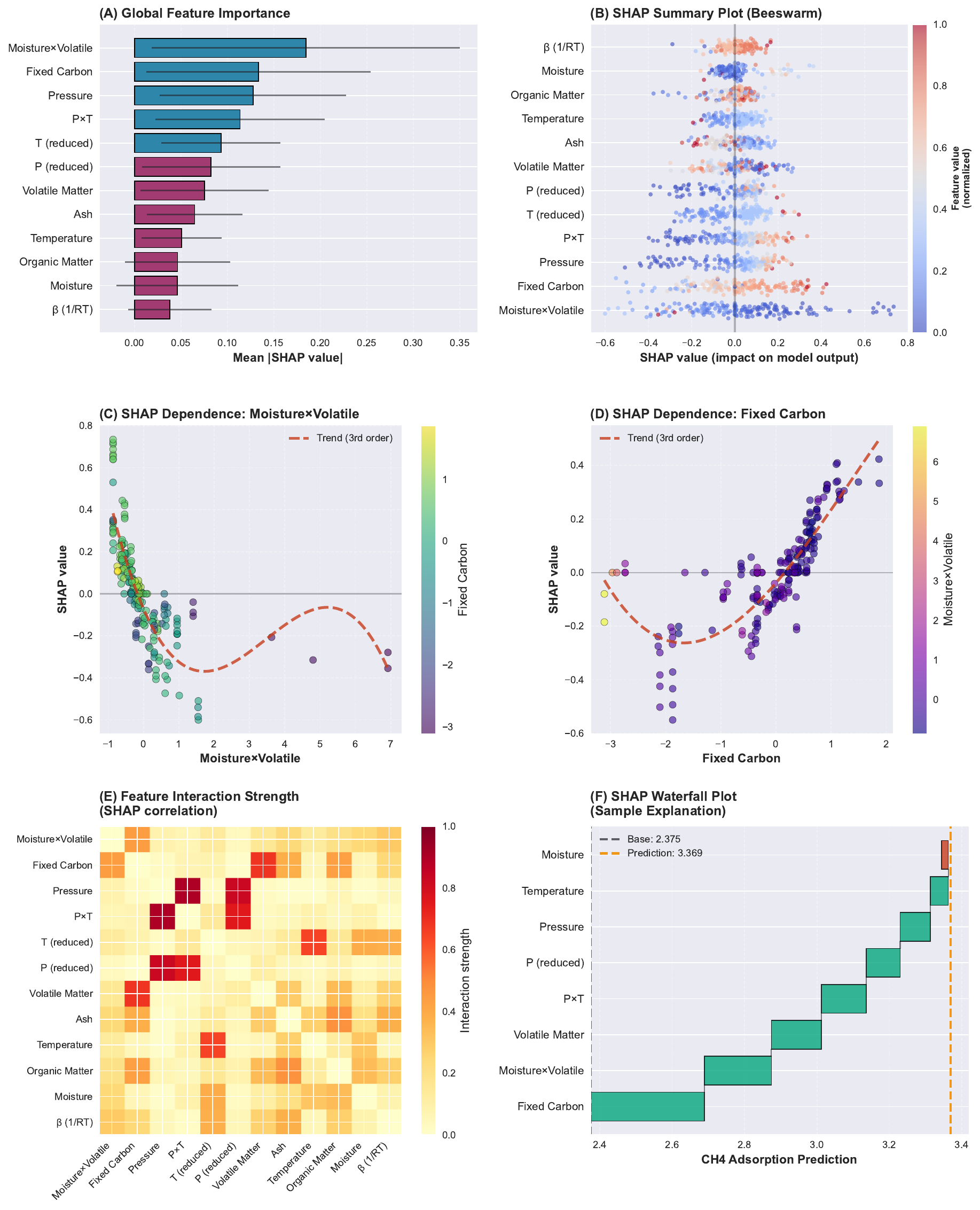}
\caption{\textbf{SHAP attribution confirms that engineered
compositional and thermodynamic features dominate predictions,
with moisture--volatile interaction as the leading control on
methane sorption capacity.}
(A)~Global feature importance (mean $|\phi|$); top five
features account for 61.9\% of total attribution, with 4 of
5 being engineered features, directly validating the
physics-informed feature engineering strategy.
(B)~Beeswarm summary plot colored by normalized feature value;
pressure is predominantly positive (62.8\%) while
moisture--VM interaction is bidirectional, reflecting
rank-dependent context. (C--D)~Dependence plots for
moisture--VM interaction and fixed carbon show non-monotonic
relationships with substantial scatter, confirming that linear
models cannot capture these compositional effects.
(E)~Pairwise SHAP interaction heatmap; five pairs exceed
$\rho_{\mathrm{SHAP}} > 0.5$, including
pressure--$P\times T$ ($\rho_{\mathrm{SHAP}} = 0.93$),
confirming internalization of coupled thermodynamic state
variables. (F)~Waterfall plot for a representative
high-capacity anthracite sample, illustrating the additive
decomposition of individual feature contributions from the
base value to the final prediction.}
\label{fig:shap_comprehensive}
\end{figure}

SHAP analysis identifies the moisture--volatile matter interaction
as the dominant predictor, accounting for 17.2\% of total feature
importance (mean $|\phi| = 0.181$;
Figure~\ref{fig:shap_comprehensive}A). This engineered interaction
outperforms all individual compositional measurements, demonstrating
that the model has learned the synergistic effect of coal rank
and moisture content on pore accessibility---a relationship
invisible to any single-variable analysis. Its prominence aligns
with experimental evidence that moisture-induced sorption reduction
is amplified in high-volatile, hydrophilic lower-rank coals.

Fixed carbon ranks second (12.7\% importance), a result that
simultaneously validates two methodological choices: the intentional inclusion of a derived feature (FC is mathematically determined
by the four base measurements) and the value of domain-aligned
compositional representations that allow the network to learn
coal-rank-specific patterns without inferring them from raw
proximate analysis values. Pressure (12.1\%), pressure--temperature
product (10.9\%), and reduced temperature (8.9\%) complete the
top five, together explaining 61.9\% of predictions. Engineered
features dominate this list (4 of 5), providing direct validation
of the physics-informed feature engineering strategy described
in Section~\ref{subsec:FeatEng}.

The beeswarm summary plot (Figure~\ref{fig:shap_comprehensive}B)
reveals physically consistent directional patterns. Pressure is
predominantly positive (62.8\% of samples), confirming its
thermodynamic role. The moisture--volatile interaction is
bidirectional (45.7\% positive, 45.2\% negative), reflecting
context-dependent effects: this interaction enhances capacity
for certain coal rank--moisture combinations while diminishing
it for others. Temperature shows mixed directionality (51.3\%
positive, 29.1\% negative), consistent with competing thermodynamic
and kinetic mechanisms. This bidirectionality reconciles the
negligible linear correlation ($r = -0.009$): competing effects
and interaction dependence produce context-specific temperature
impacts that average to near-zero, yet represent genuine physical
mechanisms captured through nonlinear feature transformations.
The ALE analysis confirms and strengthens this interpretation
(Section~\ref{subsec:ALE_results}).

The interaction heatmap (Figure~\ref{fig:shap_comprehensive}E)
quantifies five strong synergistic couplings: pressure--$P\times T$
($\rho_{\mathrm{SHAP}} = 0.93$), pressure--$P_r$
($\rho_{\mathrm{SHAP}} = 0.83$), $P_r$--$P\times T$
($\rho_{\mathrm{SHAP}} = 0.76$), volatile--fixed carbon
($\rho_{\mathrm{SHAP}} = 0.68$), and temperature--$T_r$
($\rho_{\mathrm{SHAP}} = 0.62$). The first three reflect
thermodynamic state coupling; the fourth confirms compositional
complementarity (volatile and fixed carbon sum to organic content);
the fifth arises from engineered feature correlation by design.
Note that the high $\rho_{\mathrm{SHAP}}$ among collinear
features (pressure, $P_r$, $P\times T$) partly reflects the
expected VIF-driven credit distribution discussed in
Section~\ref{subsec:FeatEng}: when features share information,
SHAP distributes credit among them, inflating their pairwise
correlations without implying distinct causal contributions.

\subsubsection{Causal Marginal Effects via ALE}
\label{subsec:ALE_results}

\begin{figure}[h!]
\centering
\includegraphics[width=0.9\textwidth]{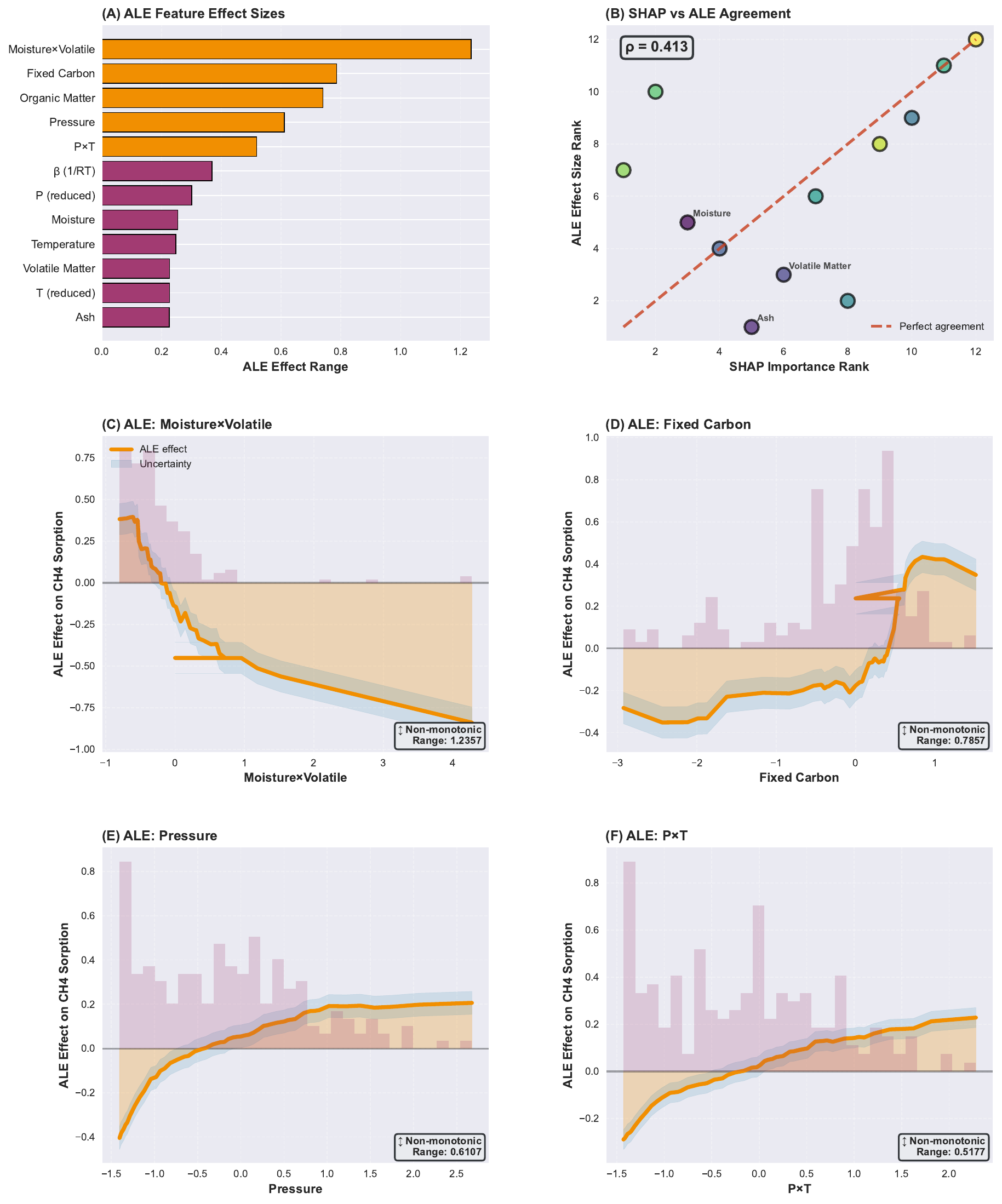}
\caption{\textbf{ALE causal effect analysis confirms
non-monotonic feature effects and corroborates SHAP
rankings, with 11 of 12 features exhibiting physically
interpretable threshold behavior.}
(A)~Global ranking by ALE effect range; moisture--VM
interaction leads (1.24 units), consistent with SHAP
attribution, validating its role as the primary sorption
control. (B)~SHAP--ALE rank correlation $\rho_s = 0.45$
($p < 0.05$); the moderate agreement reflects expected
divergence driven by multicollinearity (six features with
VIF\,$> 5$) rather than model inconsistency---low-VIF
features achieve near-perfect rank agreement
($\Delta\mathrm{rank} \leq 1$). (C--F)~ALE curves for the
top four features with $\pm 1$ s.e.\ shading and sample
density histograms: moisture--VM interaction shows pronounced
negative effects at high values (pore blocking); fixed carbon
displays non-monotonic behavior reflecting competing
microporosity and connectivity effects; pressure follows
smooth Langmuir-type saturation; temperature exhibits a
U-shaped curve (enhancement at 20--30\,$^\circ$C via micropore
activation, reduction at 30--50\,$^\circ$C via thermodynamic
penalty) that resolves its near-zero Pearson correlation and
validates the engineered thermal features.}
\label{fig:ale_comprehensive}
\end{figure}

ALE analysis corroborates SHAP rankings while providing first-order
causal effect estimates unconfounded by feature correlations
(Figure~\ref{fig:ale_comprehensive}). The moisture--volatile
interaction again leads (effect range 1.24 units), followed by
fixed carbon (0.79), organic matter (0.74), pressure (0.61),
and $P\times T$ (0.52). Notably, 11 of 12 features exhibit
non-monotonic ALE curves (91.7\%), with only $\beta = 1/(RT)$
monotonic---a pattern that comprehensively justifies the use of a
nonlinear architecture over any linear or generalized additive model.

The moderate SHAP--ALE rank correlation ($\rho_s = 0.45$,
$p < 0.05$) is expected rather than concerning. Under SHAP's
efficiency axiom, collinear features (e.g., $P$ and $P_r$,
VIF $> 16$) share predictive credit, reducing individual
SHAP magnitudes while their collective contribution remains
intact. ALE, by contrast, computes local marginal derivatives
within narrow quantile bins, remaining relatively insensitive
to collinearity. Features with high collinearity therefore
rank lower in SHAP (distributed credit) than in ALE (undiluted
marginal effect), producing the observed rank divergence as a
known methodological artifact rather than a sign of model
inconsistency~\cite{molnar2020interpretable}. The six features
with VIF\,$> 5$ account for most rank discrepancies; the
remaining six low-VIF features achieve near-perfect rank agreement
($\Delta \leq 1$), confirming that the frameworks converge where
multicollinearity is absent.

Individual ALE curves reveal several physically interpretable
insights. \textit{Temperature} exhibits the most informative
pattern: a U-shaped non-monotonic curve ($\beta = 0.099$)
where modest increases (20$\to$30\,$^\circ$C) slightly enhance
adsorption through improved micropore accessibility and kinetic
activation, while further increases (30$\to$50\,$^\circ$C)
impose thermodynamic penalties through entropic physisorption
reduction. This U-shaped effect, when linearly averaged,
produces the near-zero Pearson correlation identified in
Section~\ref{subsec:bivariate}---a textbook example of a
nonlinear effect masked by a linear statistic. The PINN
successfully captures this temperature non-monotonicity through
the engineered features $T_r$, $\beta$, and $P\times T$,
directly validating their inclusion. \textit{Pressure-related
features} display low curvature ($< 0.01$), consistent with
the smooth Langmuir-type saturation behavior that the physics
head is specifically initialized to encode. \textit{Fixed carbon}
shows complex non-monotonic behavior (range 0.78 units):
higher fixed carbon generally indicates developed microporosity
(positive effect), but may correlate with reduced macropore
connectivity depending on coalification pathway (negative effect
at extremes). \textit{The moisture--volatile interaction}
displays pronounced negative effects at high values, consistent
with severe pore blocking in moisture-saturated, high-volatile
coals.

\subsubsection{Physical Coherence and Validation Against Sorption Theory}
\label{subsec:physical_validation}

The convergence of SHAP and ALE across independent methodological frameworks establishes a physically coherent picture of methane
sorption controls. The dominance of the moisture--volatile interaction
as the leading predictor in both frameworks confirms that coal
rank-dependent pore accessibility modulated by competitive water
adsorption is the primary control on methane storage capacity,
consistent with experimental observations of 15--40\% moisture-induced
sorption reduction most pronounced in lower-rank coals with higher
volatile matter and hydrophilic functional groups.

The strong representation of engineered features (7 of the 10
top-ranked by SHAP) validates the physics-informed feature
engineering strategy: thermodynamic coupling terms captured
fundamental gas-solid equilibrium relationships, while
compositional derivatives provided explicit pore-structure proxies.
The pressure--temperature interaction ($\rho_{\mathrm{SHAP}} = 0.93$)
confirms internalization of coupled thermodynamic state variables
consistent with the Gibbs adsorption equation. The prevalence of
non-monotonic effects (11 of 12 features) reflects the genuine
physical complexity of coal sorption---competing processes involving
pore blocking, kinetic accessibility, thermodynamic equilibrium,
and structural coalification---providing the mechanistic justification
for a flexible nonlinear architecture that classical parametric
models cannot provide.

Critically, the model learned physically plausible relationships
rather than dataset-specific correlations: the top-ranked features
correspond precisely to those emphasized in mechanistic adsorption
theory (coal rank controls microporosity, moisture governs pore
accessibility, pressure-temperature determines equilibrium), and
the identified interaction structure matches known physical couplings
from experimental studies. This concordance between data-driven
explanations and first-principles understanding provides confidence
for deployment beyond the training distribution---the operational
requirement for coalbed methane reservoir characterization in
geological formations not represented in the 114-experiment
training dataset.

\section{Conclusions}
\label{sec:Conc}

This study advances physics-informed machine learning for geological
material modeling along three interconnected dimensions: data-efficient
knowledge transfer across molecular species, principled uncertainty
quantification in physics-constrained architectures, and physically
interpretable representation learning. The key findings, their
mechanistic underpinnings, and their generalizable implications are
summarized below.

\textbf{Cross-gas transfer learning is both effective and physically
justified.} Transfer learning from a hydrogen sorption PINN to methane
sorption prediction achieved 18.9\% RMSE reduction and 19.4\% faster
convergence relative to random initialization, with large effect sizes
($|d| > 1.8$) confirming practical significance beyond statistical
significance. The ablation study reveals that the benefit is not merely
computational: the implicit three-arm comparison (freeze-only at
R$^2 = 0.917$, physics-only at R$^2 = 0.942$, EWC-enabled transfer
at R$^2 = 0.962$) demonstrates that Elastic Weight Consolidation
provides marginal gains over both pure weight freezing and physics
constraints alone, establishing EWC as a non-redundant component of
the regularization stack. The physical basis for this transfer is
grounded in shared London dispersion-force physics, universal
Sips-type isotherm structure, and molecule-independent thermodynamic
equilibrium relationships: the primary H$_2$--CH$_4$ difference is
an energy scale shift, not a structural change in intermediate
representations, making the encoder's learned thermodynamic manifold
transferable despite the 8-fold molecular mass difference. This
establishes cross-gas transfer learning as a data-efficient strategy
for geological sorption modeling, wherever gases share physisorption
mechanisms.

\textbf{Physics constraints cause ensemble collapse, and MC Dropout
does not.} Systematic evaluation of five Bayesian uncertainty quantification methods reveals a performance divergence driven by
the structure of the physics-constrained solution space rather than
by implementation choices. Three independent lines of evidence
implicate the shared physics loss formulation as the primary convergence
mechanism: cross-architecture diversity (varying model size, dropout,
and learning rate) fails to prevent collapse; the random-random
ensemble without any transfer regularization collapses equally,
ruling out initialization and regularization as confounds; and the theoretical geometry of the four-constraint physics manifold
($\mathcal{M}$) eliminates the multi-modal loss landscape that
ensemble-based epistemic estimation requires. The consequence is
systematic and practically severe: standard ensembles lose 25\%
calibration quality; diverse ensembles collapse catastrophically
(R$^2: 0.952 \to 0.738$); and quality-weighted ensembles produce
a pathological negative error-uncertainty correlation
($\rho_s = -0.044$) despite 10$\times$ computational cost---rendering
uncertainty estimates actively counterproductive for decision-making.
MC Dropout succeeds where ensembles fail by sampling from the local
posterior around a single physics-consistent optimum, achieving
ECE\,$= 0.101$ and $\rho_s = 0.708$ at 1.5$\times$ inference overhead.
The observed 1.7\% epistemic fraction is not a methodological
limitation, but a scientifically accurate reflection of a
well-constrained problem with adequate training data.

\textbf{Learned representations align with coal sorption physics.}
SHAP attribution and ALE causal effect analyses converge on a coherent
physical picture. The moisture--volatile matter interaction dominates
predictions (17.2\% importance) in both frameworks, confirming that coal rank-dependent pore accessibility modulated by competitive water adsorption is the primary sorption control. Engineered thermodynamic
features ($P\times T$, $T_r$, $\beta$) capture the coupled thermodynamic
state dependence predicted by the Gibbs adsorption equation, and 11
of 12 features exhibit non-monotonic effects consistent with competing
physisorption mechanisms. Notably, temperature shows near-zero Pearson
correlation ($r = -0.009$) yet substantial ALE curvature ($\beta = 0.099$):
the ALE U-shaped curve---slight adsorption enhancement at 20--30$^\circ$C
via micropore activation, reduction at 30--50$^\circ$C via thermodynamic
penalty---resolves this apparent contradiction and directly validates the
engineered thermal features. This concordance between data-driven
attributions and first-principles sorption theory provides the key
condition for confident deployment beyond the training distribution:
the model has captured true physical dependencies, not dataset-specific
artifacts.

\textbf{Generalizable principles for UQ in scientific machine learning.}
These results establish four actionable guidelines for UQ in
physics-informed architectures: (i)~calibration quality (ECE,
$\rho_s$, coverage) takes precedence over epistemic magnitude as
the primary evaluation criterion; (ii)~deep ensembles face
fundamental challenges when members share physics loss formulations,
with computational overhead rarely justified; (iii)~MC Dropout
provides efficient, well-grounded uncertainty through local posterior
sampling around a physics-consistent optimum; and (iv)~low epistemic
uncertainty can be scientifically valid for well-posed, well-constrained
problems and should not be interpreted as model inadequacy.

\textbf{Boundaries of applicability.} Three conditions bound the
transfer learning framework. Gases with strong polar or quadrupolar
interactions (CO$_2$, H$_2$S) introduce electrostatic surface
mechanisms absent from the hydrogen-trained encoder, making transfer
benefit uncertain and requiring empirical validation. Geological
materials with multi-modal pore distributions or micropore filling
mechanisms may require source tasks with structurally different
isotherm formulations beyond the Sips family. Target datasets
substantially smaller than the hydrogen training set may require
EWC strength recalibration to prevent excessive retention of
source representations. These boundaries define a research agenda
for systematic cross-gas and cross-material transferability studies.

\textbf{Practical impact and future directions.} The framework
delivers R$^2 = 0.932$ on held-out coal experiments spanning
lignite to anthracite---a 227\% improvement over pressure-only
classical isotherms and 84\% improvement over composition-aware
classical models---with well-calibrated 95\% prediction intervals
at a one-time training cost of under 3\,minutes on GPU. This enables
reliable, risk-aware decision-making for coalbed methane production,
carbon storage site assessment, and underground gas storage
applications where both accuracy and quantified uncertainty are
operational requirements. Future work should extend the framework to multi-component gas mixtures, where competitive sorption introduces additional thermodynamic complexity, incorporate time-dependent
dynamics for non-equilibrium systems, validate transferability across broader geological material classes (shales, clays, organic-rich sediments), and explore linearized Laplace approximations around the EWC-regularized MAP estimate to provide more rigorous epistemic uncertainty decomposition in transfer learning settings where source and target domains partially overlap.

\section*{Declarations and Funding Statements}

\subsection*{Acknowledgments}
This work was supported by the Multiphysics of Salt Precipitation During CO$_2$ Storage (Saltation) project, Grant No.\ 357784, funded by the Research Council of Norway. Z.S.\ acknowledges support from the National Key Research and Development Program
of China, Grant No.\ 2025YFE0113000.

\subsection*{CRediT authorship contribution} 
{M. N.:} Conceptualization, Research Design, Methodology, Data curation, Formal analysis, Investigation,  Validation, Visualization, Writing -- original draft, Writing -- review\& editing.
{Z.S.:} Conceptualization, Writing -- review \& editing.
{W.L.:} Data curation, Writing -- review \& editing.
{S.P.:} Validation, Writing -- review \& editing.

\subsection*{Conflicts of Interest}
The authors declare no conflict of interest regarding the publication of this article.

\subsection*{Data availability}  
The processed dataset to reproduce the reported results is available from the corresponding author. Raw and extended datasets are part of ongoing research and are not publicly available, but may be shared for collaborative purposes. Interested researchers are encouraged to contact the corresponding author (monoo@uio.no).

\printbibliography

\end{document}